\documentclass{article}

\usepackage[dandb, final]{neurips_2025}

\usepackage{makecell} 
\usepackage{graphicx}
\usepackage{booktabs} 
\usepackage[table]{xcolor}
\usepackage[backref=page]{hyperref} 
\hypersetup{
    colorlinks,
    linkcolor={OrangeRed3},
    citecolor={SpringGreen4},
    urlcolor={PaleTurquoise4}
}
\usepackage{array}
\usepackage{graphicx}
\usepackage{subcaption}
\usepackage{multirow}
\usepackage[dvipsnames,svgnames,x11names]{xcolor}
\usepackage{wrapfig}
\usepackage{amsmath}

\usepackage[utf8]{inputenc} 
\usepackage[T1]{fontenc}    
\usepackage{hyperref}       
\usepackage{url}            
\usepackage{booktabs}       
\usepackage{amsfonts}       
\usepackage{nicefrac}       
\usepackage{microtype}      
\usepackage{xcolor}         
\usepackage{graphicx}  

\title{Open-Insect: Benchmarking Open-Set Recognition of Novel Species in  Biodiversity Monitoring}

\author{
  \textbf{Yuyan Chen\textsuperscript{1,2}},
  \textbf{ Nico Lang\textsuperscript{3}},
  \textbf{ B. Christian Schmidt\textsuperscript{4}},
  \textbf{ Aditya Jain\textsuperscript{2}},\\
  \textbf{ Yves Basset\textsuperscript{5,6,7}},
    \textbf{ Sara Beery\textsuperscript{8}},
      \textbf{ Maxim Larrivée\textsuperscript{9}},
        \textbf{ David Rolnick\textsuperscript{1,2}},
  \\
  \textsuperscript{1}McGill University \quad
  \textsuperscript{2}Mila - Quebec Artificial Intelligence Institute \quad 
  \textsuperscript{3}University of Copenhagen \quad \\ 
  \textsuperscript{4}Agriculture and Agri-food Canada \quad 
  \textsuperscript{5}Smithsonian Tropical Research Institute \quad \\
  \textsuperscript{6}Biology Center, Czech Academy of Sciences \quad 
  \textsuperscript{7}Maestria de Entomologia, University of Panama \quad \\
  \textsuperscript{8}Massachusetts Institute of Technology \quad  
  \textsuperscript{9}Montréal Insectarium \quad
}

\begin{document}

\maketitle

\begin{abstract}
Global biodiversity is declining at an unprecedented rate, yet little information is known about most species and how their populations are changing. Indeed, some 90\% of Earth's species are estimated to be completely unknown. Machine learning has recently emerged as a promising tool to facilitate long-term, large-scale biodiversity monitoring, including algorithms for fine-grained classification of species from images. However, such algorithms typically are \emph{not} designed to detect examples from categories unseen during training -- the problem of open-set recognition (OSR) -- limiting their applicability for highly diverse, poorly studied taxa such as insects. To address this gap, we introduce Open-Insect, a large-scale, fine-grained dataset to evaluate unknown species detection across different geographic regions with varying difficulty.
We benchmark 38 OSR algorithms across three categories: post-hoc, training-time regularization, and training with auxiliary data, finding that simple post-hoc approaches remain a strong baseline. We also demonstrate how to leverage auxiliary data to improve species discovery in regions with limited data. Our results provide insights to guide the development of computer vision methods for biodiversity monitoring and species discovery. 

\end{abstract}

\begin{figure}[!htb]
\centering
\begin{center}
\small
\includegraphics[width=0.95\columnwidth]{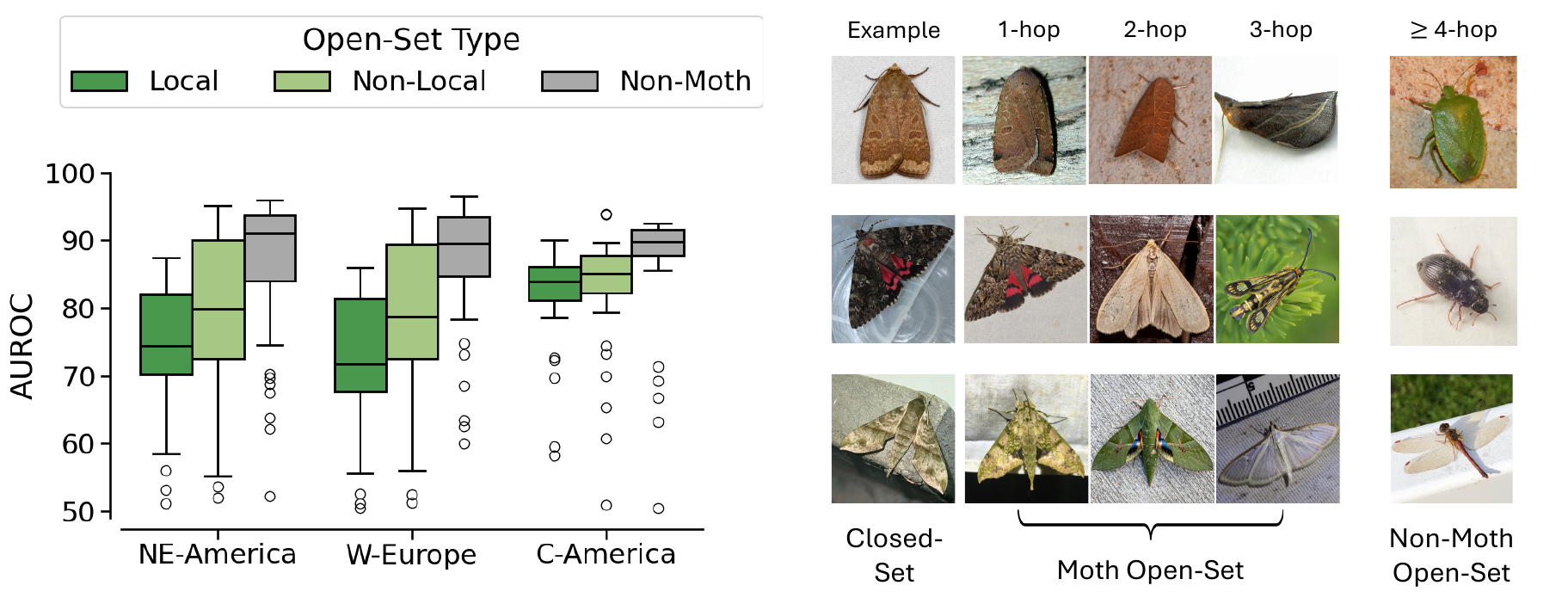}
\caption{\textbf{Open-Insect benchmark results on three geographical regions with varying difficulty.} The Open-Insect benchmark includes images of thousands of highly visually similar moth species, along with non-moth arthropods, divided by geographic region. \textbf{Left}: Results from 38 OSR methods on three open-set types i) \emph{Local} moth, ii) \emph{Non-local} moth, and iii) \emph{Non-moth} (see Table~\ref{tab: main result}). \textbf{Right}: Visual dissimilarity across taxonomic levels: 1-hop (same genus), 2-hop (different genus, same family), 3-hop (different family within \emph{Lepidoptera}), and non-moths (different order, $\geq$4 hops). }
\label{fig:teaser_boxplot_examples}
\end{center}
\vskip -2.0em
\end{figure}
\section{Introduction}

The use of machine learning (ML) for species recognition can greatly facilitate long-term, large-scale biodiversity conservation \cite{tuia2022perspectives}. In recent years,  significant progress has been made in developing sophisticated ML and computer vision tools in biodiversity, including a variety of challenging benchmarks \cite{van2018inaturalist, van2021benchmarking, vendrow2024inquire, vcermak2024wildlifedatasets} and foundation models \cite{NEURIPS2023_87dbbdc3,gharaee2024bioscan, stevens2024bioclip, yang2024biotrove}, which span hundreds of millions of images and hundreds of thousands of species. 

Most ML models are trained with the closed-world assumption, assuming that at inference time, the model will only encounter categories that are seen during training. However, in biodiversity monitoring, the closed-world assumption is often violated for several reasons. Firstly, an estimated 86 percent of terrestrial species and 91 percent of marine species remain undescribed \cite{mora2011many}. Regardless of the size and diversity of the training dataset, unknown species can readily occur when deploying a model in practice, especially with less well-studied groups such as insects or deep-sea fauna. Secondly, not all described species have data available for training. For example, there are over 2 million species according to the latest Catalogue of Life checklist \cite{Banki2025}, but only 65 percent of them have a record on the Global Biodiversity Information Facility (GBIF)\footnote{\href{https://www.gbif.org/}{https://www.gbif.org/} (accessed 2025-01-29)}, which compiles records from across biodiversity platforms such as iNaturalist\footnote{\href{https://www.inaturalist.org/}{https://www.inaturalist.org/} (accessed 2025-01-29)}. Thirdly, very often, species recognition models are trained with regional checklists of species that are of interest in specific use cases. Such models cannot correctly identify species that are out of scope even if they are well documented. 

For these reasons, open-set recognition (OSR), which aims at accurately classifying closed-set samples (samples that belong to a category seen during training) and detecting open-set (OS) samples, should be a crucial component of ML for biodiversity monitoring. Beyond improving the accuracy of existing identification systems, OSR algorithms can help to pinpoint previously undiscovered species and invasive or introduced species in specific geographic regions. Most existing OSR benchmarks \cite{yang2023imagenet, galil2023framework, hendrycks2021natural, bitterwolf2023or} are derived from ImageNet-1K \cite{russakovsky2015imagenet} and ImageNet-21K \cite{deng2009imagenet}, which do not capture the complexities and fine-grained categories of biodiversity data \cite{luccioni2023bugs} or its long-tailed distribution. While some OSR benchmarks do contain biodiversity categories, they are relatively small in scale and focus on well-studied taxa such as birds \cite{vaze2021open} and other vertebrates \cite{hogeweg2024cood}. As a result, performance on these benchmarks may not accurately represent performance in new species detection for poorly documented, highly diverse taxa.

An ideal benchmark dataset for new species detection should contain taxa with plentiful undiscovered species. Since about two-thirds of all animals are insects \cite{Banki2025} and over 80 percent of insect species remain undescribed \cite{stork2018many}, we introduce Open-Insect, a large-scale, fine-grained image dataset which focuses on insects (see Fig.~\ref{fig:teaser_boxplot_examples} for examples). Open-Insect consists of closed-, open-set, and auxiliary splits for three geographical regions: Northeastern North America, Western Europe, and Central America. 
Following Tobler's first law of geography\footnote{``\textit{Everything is related to everything else, but near things are more related than distant things.}''}, we hypothesize that the difficulty of detecting semantic shifts correlates with geographic proximity. Therefore, Open-Insect utilizes geographical metadata to study local and non-local semantic shifts. Local open-set species may not only be harder to detect, but also reflect challenges encountered when deploying species-recognition models in practice. For each region, we also include a realistic auxiliary dataset for OSR methods that benefit from training with such data. 

Our contributions can be summarized as follows:
\begin{enumerate}
\setlength{\itemsep}{0em}
  \item We introduce Open-Insect, a large-scale, extremely fine-grained biodiversity dataset\footnote{Dataset available at: \href{https://huggingface.co/datasets/yuyan-chen/open-insect}{https://huggingface.co/datasets/yuyan-chen/open-insect}} focused on insects for open-set recognition. Open-Insect allows us to benchmark the performance of 36 existing OSR methods for species discovery and invasive species detection\footnote{Code available at: \href{https://github.com/RolnickLab/Open-Insect}{https://github.com/RolnickLab/Open-Insect}}. 
    \item We show that the quality of auxiliary data is crucial to improve OSR performance, and present a simple approach to benefit from auxiliary training data. 
    \item We observe that the maximum softmax probability (MSP) and methods derived from it remain a strong baseline for fine-grained OSR. 
\end{enumerate}

\section{Related Work}

\subsection{OSR and OOD Detection}

The goal of OSR is to accurately classify the closed-set categories and detect the open-set ones \cite{scheirer2012toward}. OpenMax \cite{bendale2016towards} is the first deep learning-based method for open-set recognition. More recent methods include ARPL \cite{chen2021adversarial}, OpenGAN \cite{kong2021opengan}, and hierarchy-adversarial learning \cite{lang2024coarse}. Out-of-distribution (OOD) detection is a broader term that refers to any setting in which one must detect test samples drawn from a different distribution than the training distribution. The shift can be either covariate or semantic. Covariate shift occurs when OOD samples come from a different input space, while semantic shift arises when new categories (labels) occur in the test set \cite{yang2024generalized}. The majority of recent OOD detection methods focus on semantic shift and tackle the same problem as OSR \cite{yang2024generalized}, making them suitable for species discovery as well. Hence, we also compare against these OOD detection methods in our work.

We categorize the aforementioned OOD detection methods into 1) post-hoc, 2) training-time regularization, and 3) training with auxiliary data. \emph{Post-hoc} methods, as the name suggests, do not require any further training and can be easily integrated into an existing pipeline. These methods utilize the outputs of an already trained classifier and develop a score function to map the outputs to a real number \cite{bendale2016towards, hendrycks2016baseline, guo2017calibration, liang2017enhancing, lee2018simple,lee2018simple, ren2021simple, pmlr-v119-sastry20a, liu2020energy, kong2021opengan, huang2021importance, sun2021react, hendrycks2019scaling, vaze2021open, pmlr-v162-hendrycks22a, wang2022vim, sun2022out, sun2022dice, song2022rankfeat, djurisic2022extremely, zhang2022out, ammar2023neco, liu2023fast, jiang2023detecting, liu2023detecting}. These outputs can be, for instance, features \cite{wang2022vim, ren2021simple, lee2018simple}, logits \cite{hendrycks2019scaling}, gradients \cite{huang2021importance}, or softmax probability \cite{hendrycks2016baseline}. 
For methods that require further training, a number of methods do not require extra training data, but improve OSR performance by \emph{regularization} with new loss functions or model architectures, such as
\cite{devries2018learning, wei2022mitigating, chen2021adversarial, hsu2020generalized, hendrycks2019using, ming2022exploit, huang2021mos, du2022vos,  kong2021opengan}. However, OSR can benefit from extra data if they are available. Early works that explored the usefulness of auxiliary data include Entropic Open-Set Loss \cite{dhamija2018reducing} and Outlier Exposure \cite{hendrycks2018deep} . Recently, an increasing number of methods have been developed which involve \emph{training models with auxiliary data} \cite{hofmann2024energy, miao2024long, yao2024out, sharifi2025gradient, xu2024out, fan2024test, du2024does, jiang2023diverse, narasimhan2023plugin, bai2023feed, bai2024aha}, utilizing  either labeled or unlabeled data. 

\subsection{Benchmarks and Datasets}
\label{sec: related work benchmarks}
OSR tasks usually involve splitting categories of a dataset into the closed and the open set. On the other hand, OOD detection tasks typically involve taking an entire dataset, such as ImageNet \cite{russakovsky2015imagenet}, as in-distribution (ID) and several other datasets as OOD, such as Texture \cite{kylberg2011kylberg} or iNaturalist \cite{van2018inaturalist}. OpenOOD \cite{yang2022openood} is the first benchmark to standardize the evaluation of OOD detection and OSR methods. The 4 OOD benchmarks follow the conventional setup as mentioned above. The 4 OSR benchmarks were constructed by splitting small-scale datasets, namely MNIST \cite{deng2012mnist}, CIFAR10 \cite{krizhevsky2009learning}, CIFAR100 \cite{krizhevsky2009learning}, and TinyImageNet-200 \cite{torralba200880} into the closed and the open sets. A series of datasets has also been developed in the same way by splitting  ImageNet-1K \cite{russakovsky2015imagenet} and ImageNet-21K \cite{deng2009imagenet} to evaluate semantic shift detection while minimizing covariate shifts, including ImageNet-OOD \cite{yang2023imagenet}, COOD \cite{galil2023framework}, ImageNet-O \cite{hendrycks2021natural}, and OpenImageNet \cite{wang2022vim}. OpenOOD v1.0 \cite{yang2022openood} and v1.5 \cite{zhang2023openood} are the only benchmarks, to our knowledge, to include an auxiliary dataset, TinyImageNet \cite{torralba200880}, for the 4 OOD datasets. In many works \cite{hofmann2024energy, miao2024long, bai2024aha, yao2024out, sharifi2025gradient, xu2024out, fan2024test, du2024does, jiang2023diverse, narasimhan2023plugin, bai2023feed}, training with auxiliary data methods are evaluated with different ID, OOD, and auxiliary data splits, making a fair comparison of  these works challenging. 

The Semantic Shift Benchmark (SSB) \cite{vaze2021open},  Combined Out-of-Distribution Detection (COOD) \cite{hogeweg2024cood}, and iNat21-OSR \cite{lang2024coarse} include OSR or OOD detection datasets that consist of biodiversity categories. SSB consists of an OSR dataset based on Caltech-UCSD Birds-200-2011(CUB) \cite{wah2011caltech}. COOD \cite{hogeweg2024cood} used Norwegian vertebrates \cite{schermer2018supporting} as the ID set and Non-Norwegian vertebrates \cite{schermer2018supporting} as the OOD set to evaluate OSR in hierarchical classification. Both SSB and COOD are relatively small-scale and focus on well-studied groups of animals. iNat21-OSR \cite{lang2024coarse} was developed to compare OSR performance at different semantic distances from coarse to fine-grained in the taxonomic hierarchy, from the same genus to different kingdoms. None of these datasets includes auxiliary training data.

\subsection{ML for Biodiversity Monitoring}

Machine learning is increasingly used in species recognition to support biodiversity monitoring. A number of large-scale, fine-grained species recognition datasets have been introduced, such as iNat18 \cite{van2018inaturalist}, iNat21 \cite{van2021benchmarking}, Tree-of-Life 10M \cite{stevens2024bioclip}, and BioTrove \cite{yang2024biotrove}. Recently, foundation models such as BioCLIP \cite{stevens2024bioclip} and BioTrove-CLIP \cite{yang2024biotrove} have been developed to achieve generalization across datasets, domains, and tasks, with insect-specific models like CLIBD \cite{gong2024clibd}, trained on BIOSCAN-1M \cite{NEURIPS2023_87dbbdc3} and BIOSCAN-5M \cite{gharaee2024bioscan}, and Insect-Foundation \cite{nguyen2024insect}, trained on Insect-1M, marking significant progress in this area. A detailed comparison between the insect datasets and Open-Insect is provided in the Appendix. Some of these models  achieve zero-shot classification of unseen species, by leveraging prior knowledge of these species such as scientific names, reference images, or DNA barcodes. This technique is not applicable to flagging undescribed species, which lack such prior information, but can be useful for identifying rare species or supporting checklists of species of interest.

\section{The Open-Insect Dataset} 
\begin{figure}[]
\begin{center}
\centerline{\includegraphics[width=0.95\linewidth]{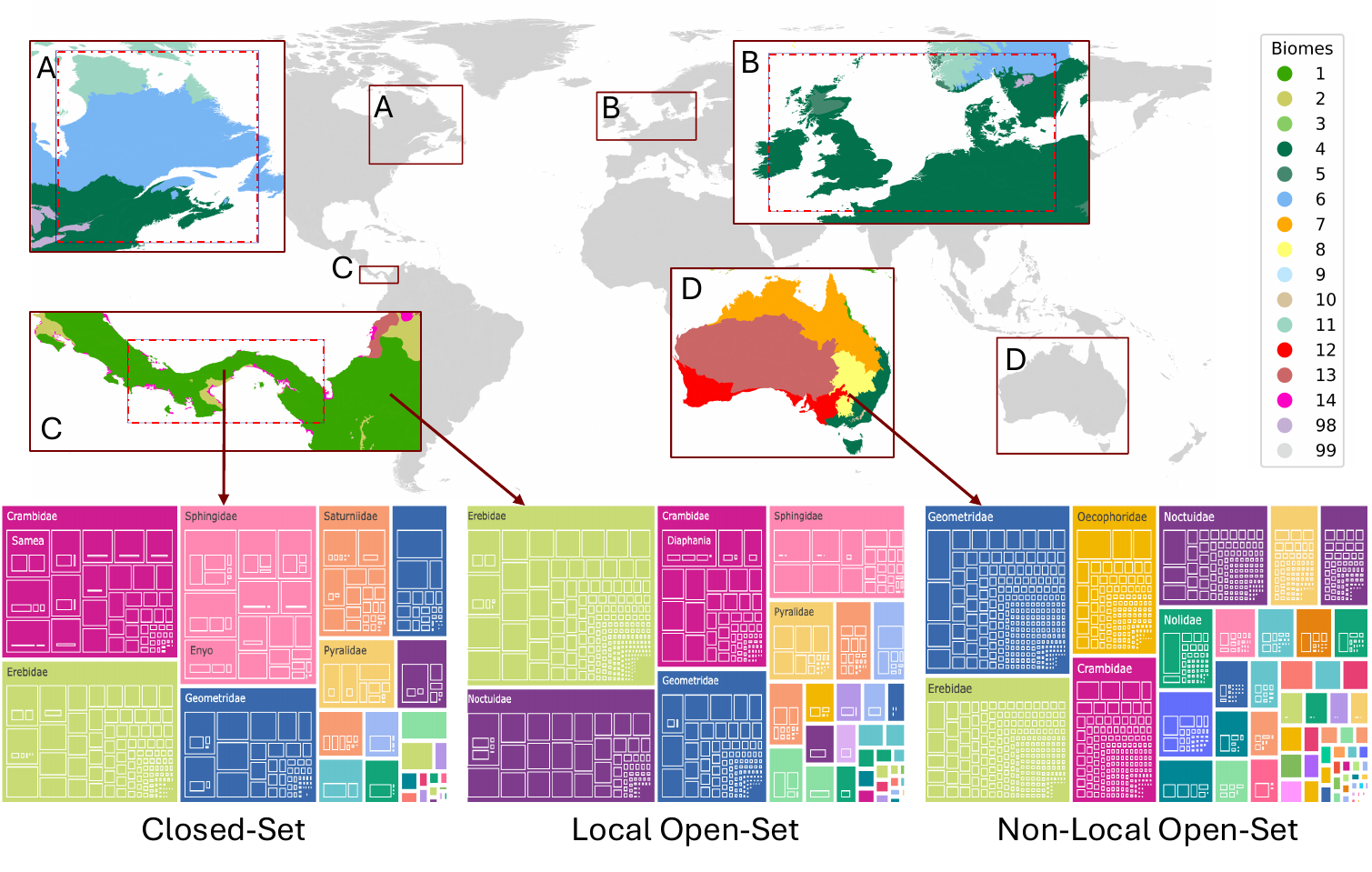}}
\caption{\textbf{Open-Insect dataset overview.} Regions A, B, and C correspond to closed-sets and local open-sets, while region D corresponds to the non-local open-set. (Region A: NE-America; B: W-Europe; C: C-America; D: Australia.) The tree maps visualize the taxonomic distribution of moth families in regions C and D, where nested boxes denote genera and species, and box size indicates the relative number of images. The same family is \emph{colored consistently} across the three treemaps. Local open-set species are more similar to the closed-set than non-local ones. Biome codes and tree maps for the other two regions are provided in the Appendix.}
\label{fig: global biome map}
\end{center}
 \vskip -2em
\end{figure}

Of all insects, we focus on moths in this work, as moths are easily attracted to ultraviolet  camera traps \cite{bjerge2021automated} and many can be visually identified to the species level, allowing automatic large-scale long-term monitoring \cite{roy2024towards, van2022emerging, jain2024insect}. These monitoring systems are expected to encounter a plethora of novelties \cite{ronkay2018agrotis, sondhi2021moths}, making it especially timely to develop ML-based models that can accommodate undocumented species and accelerate species discovery. We curate Open-Insect based on the AMI dataset \cite{jain2024insect}, a large-scale, fine-grained dataset consisting of 5,364 moth species and 12 groups of non-moth arthropods in three regions: Northeastern North America (NE-America), Western Europe (W-Europe), and Central America (C-America) (see Fig.~\ref{fig: global biome map}: areas inside the dashed bounding boxes A, B, and C). We summarize the Open-Insect data distribution in Table \ref{tab: the Open-Insect dataset}.

\subsection{The Closed-Set}
\label{sec: ID data}
Our benchmark dataset uses the AMI-GBIF dataset \cite{jain2024insect} as in-distribution data. We follow the regional splits in the AMI dataset, as the first two regions represent well-documented regions, while the last represents regions with very high biodiversity but limited data. Beside the AMI-GBIF dataset, we compiled 2,912,168 images of 28,388  moth species from GBIF to minimize covariate shift for open-set and auxiliary species (see the Appendix for curation details).

\subsection{The Open-Sets}
We designed three open-sets: 1) Local moth (O-L), 2) Non-Local moth (O-NL), 3) and Non-Moth (O-NM), to simulate different kinds of unseen species that a model may encounter in the wild -- respectively: 1) species related to those known by the model but which are undocumented (no data or new to science), 2) introduced or invasive species that have only recently colonized the region, and 3) species that are not the focus of the model, but may occur in the monitoring system.

\textbf{Local moths.} Since we have very limited data on truly new species, we need to use described species to simulate them. To gather a set of local open-set species, we extended each region by 1 degree in latitude and 3 degrees in longitude, and then compiled a checklist containing species that have at least one occurrence in the extended area (Fig. \ref{fig: global biome map}: areas inside the solid bounding boxes A, B, and C) and excluded all species already present in the closed-set. Since all three regions share similar biomes with their neighboring areas (see Fig.~\ref{fig: global biome map}), it is likely that many of these local open-set species are also present but yet to be observed in the given region, making them ideal substitutes for new species. 

\begin{wrapfigure}{r}{0.4\textwidth}
\begin{center}
\centerline{\includegraphics[width=0.38\columnwidth]{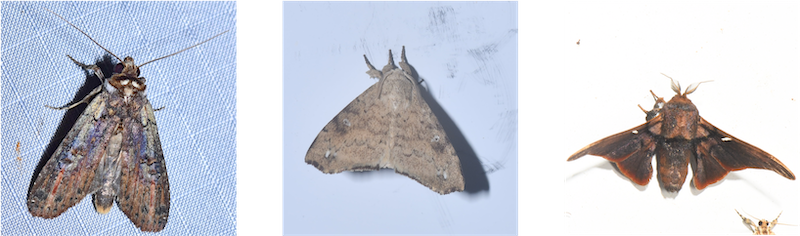}}
\caption{\textbf{Examples of likely novel species from BCI.} These species show greater than 7\% DNA barcode divergence from their closest match in the BOLD database, well above the ~1.5\% species-level cutoff.}
\label{fig: bci_new_species_image_only}
\end{center}
\vskip -1.5em
\end{wrapfigure}

For C-America, we also include 197 images recently collected from Barro Colorado Island (BCI), Panama, gathered by our team to provide a maximally realistic test case. This dataset split, referred to as \textbf{O-B} in Table \ref{tab: the Open-Insect dataset}, includes 133 open-set species—59 of which are likely undescribed by science (see examples in Fig.~\ref{fig: bci_new_species_image_only}). Though these species have not gone through the formal accreditation process yet, they exhibit over 1.5\% DNA sequence divergence from their closest match in the Barcode of Life Data (BOLD) database \cite{ratnasingham2007bold}. This is a commonly used threshold for distinguishing similar but distinct sibling species \cite{zahiri2017probing, ratnasingham2013dna}.  The O-B split was not used for the benchmark results reported in Table \ref{tab: main result}, but to assess 1) whether methods that perform well on Open-Insect also generalize well to potentially new species, and 2) whether the OSR score threshold, determined by fixing the false positive rate on Open-Insect, remains effective for species discovery in real-world scenarios.

\textbf{Non-local moths.} Here we simulated non-native species (such as introduced invasive species) by selecting all species occurring in Australia (Fig. \ref{fig: global biome map}: region D) and falling outside the closed-sets and local open-sets. Australia was picked since it is geographically isolated and far from all closed-set regions. This set contains 4,662 species and 95,597 samples. In Fig. \ref{fig:teaser_boxplot_examples},  we show the visual dissimilarity of moth species at different taxonomic distances: 1-hop (from the same genus), 2-hop (from a different genus but same family), and 3-hop  (from a different family). The detailed comparison of the local and non-local open-sets are summarized in the Appendix. In short, local open-sets have substantially more 1-hop species than non-local. 

\textbf{Non-moths.} To evaluate ML algorithms' performance on rejecting species that are beyond the scope of study, but may occur in monitoring sites, we randomly sampled 35,000 images of non-moth species from the AMI-GBIF dataset. These species are mostly other insects, along with some arachnids. Examples are shown in Fig.  \ref{fig:teaser_boxplot_examples}.

\subsection{Auxiliary Data}
As mentioned in Sec. \ref{sec: related work benchmarks}, when comparing OSR algorithms that use auxiliary data, it is important that such data be consistent across methods. This motivates us to explicitly include an auxiliary training set for each region in the Open-Insect benchmark, in contrast to previous benchmarks. By including auxiliary datasets, our goal is not merely to expand the training set, but to curate a \emph{unified} and \emph{comprehensive} benchmark to facilitate the development and evaluation of OSR methods that were not previously evaluated with biodiversity-related datasets \cite{vaze2021open, hogeweg2024cood, lang2024coarse}.

For each regional dataset, we constructed the auxiliary set (AUX) as follows. We first excluded all species that are in the closed-set, the local open-set, or the non-local open-set. Since species within the same genus can be highly similar, we also excluded species from any genus present in the open-set but not the closed-set, since it is possible that a model could learn to simply declare all species from this genus as unknown, thereby inflating performance metrics. For NE America and W Europe, we selected 8,000 species for training, using 20 images per species. For Central America, we selected 4,000 species, each with 20 images. We also randomly sampled a small set of images from the auxiliary set to construct regional validation sets for hyperparameter tuning of the post-hoc methods. Details can be found in Table \ref{tab: the Open-Insect dataset}.

\begin{wraptable}{r}{0.55\textwidth}
\vskip -1em
\caption{\textbf{The Open-Insect dataset.} For each region,  we consider species in AMI-GBIF as the closed-set, and species from nearby regions as local open-set. We also provide auxiliary data with other species that are available at train time.
In addition, the Open-Insect dataset provides a non-local  open-set (O-NL) as well as a non-moth open-set (O-NM).
Species in the closed-, O-L, or AUX sets may overlap across the three regions, but not within a region. These images were compiled from public data on GBIF. For C-America, we also include 197 images of 57 potentially new species (denoted as ``O-B'') collected by our team in BCI.}
\label{tab: the Open-Insect dataset}
\begin{center}
\small 
\resizebox{0.54\textwidth}{!}{%
\begin{tabular}{ccrrrr}
\toprule
\textbf{Region} & \textbf{Type} & \makecell{\textbf{Moth} \\ \textbf{species}} &
\makecell{\textbf{Train} \\ \textbf{images}} & 
\makecell{\textbf{Val} \\ \textbf{images}} &
\makecell{\textbf{Test} \\ \textbf{images}}\\
\midrule
\multirow{3}{*}{NE-A} & Closed &2,497 &870,336 & 102,987& 206,620\\
& O-L & 617 &-&- & 113,634 \\
& AUX & 8,000 & 160,000 & 19,439 & - \\
\midrule
\multirow{3}{*}{W-E} & Closed &2,603 &1,177,125 & 134,095& 268,113  \\
& O-L & 458 &- & -& 78,648  \\
& AUX & 8,000 & 160,000 & 21,868 & -  \\
\midrule
\multirow{3}{*}{C-A} & Closed &636  &72,188 & 9,054& 18,163 \\
& O-L & 1,724  &-&-& 111,879  \\
& O-BCI & 133 & - & - &  197 \\
& AUX &4,000 & 80,000 &9,054 & - \\
\midrule
 \multirow{2}{*}{\makecell{-}} &  O-NL  & 4,662 &-&-& 95,597  \\
  &  O-NM & -  &-&-& 35,000 \\
\bottomrule
\end{tabular}
}
\end{center}
\vskip -1.2em
\end{wraptable}

\section{Methods}
\label{sec: methods}
 \textbf{Existing methods}.
 Similar to OpenOOD v1.5 \cite{zhang2023openood}, we divided the methods into three categories: 1) post-hoc; 2) training-time regularization; and 3) training with auxiliary data. For post-hoc methods, we evaluate all such methods in OpenOOD v1.5 \cite{zhang2023openood} and also more recent methods, including NECO \cite{ammar2023neco}, FDBD \cite{liu2023fast}, RP \cite{jiang2023detecting}, and NCI \cite{liu2023detecting}.  For training-time regularization, we evaluate ConfBranch \cite{devries2018learning}, G-ODIN \cite{hsu2020generalized}, ARPL \cite{chen2021adversarial}, LogitNorm \cite{wei2022mitigating}, and RotPred \cite{hendrycks2019using}. Though OpenGAN \cite{kong2021opengan} is considered as post-hoc in OpenOOD v1.5 \cite{zhang2023openood}, we followed OpenOOD \cite{yang2022openood} and categorize it as ``training-time regularization'' because it still requires training the generator and the discriminator with features extracted by the classifier, making it more complicated to use than other post-hoc methods. We evaluated all ``training with extra data'' methods in OpenOOD v1.5 \cite{zhang2023openood} except for MCD \cite{yu2019unsupervised}, due to its extremely low closed-set classification accuracy, and included Energy \cite{liu2020energy}.

\textbf{Proposed baselines for training with auxiliary data}.
 We additionally propose two simple baselines that utilize auxiliary data. Let $C$ be the number of species in the closed-set and $A$ the number of species in the auxiliary set. We preserve the original $C$-dimensional classification head and simultaneously train a second linear classification head with either $C+1$ or $C+A$ dimensions -- i.e., treating the auxiliary set either as one new class or as $A$ new classes. We call these two methods \emph{NovelBranch} and \emph{Extended}, respectively. In our experiments, the models were optimized with the cross-entropy of both classification problems sharing the feature extractor. At test time, the OSR scores were computed from the closed-set head so that ID accuracy is preserved.

\textbf{Implementation details.} The implementation follows OpenOOD\footnote{\href{https://github.com/Jingkang50/OpenOOD/tree/main}{https://github.com/Jingkang50/OpenOOD/tree/main} (accessed 2025-01-29)} \cite{yang2022openood}. We used a ResNet-50 backbone \cite{he2016deep} for all methods, following a standard practice in the OSR \cite{lang2024coarse}, OOD detection \cite{yang2022openood, zhang2023openood}, and species recognition literature \cite{van2021benchmarking}. We used the AdamW optimizer \cite{loshchilov2017decoupled} and applied cosine scheduling \cite{he2019bag} to the learning rate. The weight decay was constant ($10^{-4}$) during training. All basic classifiers were trained from scratch for 120 epochs, with 0.01 initial learning rate (lr) and 6 warm-up epochs. For all training methods (with or without auxiliary data), we fine-tuned species-level classifiers for 30 additional epochs, with 0.001 initial lr and 2 warm-up epochs. We trained three basic classifiers per region and report the mean and standard deviation of their performance in Table \ref{tab: main result}. The models for training methods were fine-tuned using one randomly selected basic classifier. More training details can be found in the Appendix.

\textbf{Using pre-trained weights.} It has been shown that ImageNet pre-trained weights can improve accuracy on species recognition when the training set is small \cite{van2021benchmarking}. Since the C-America closed-set is much smaller than those for the other two regions, we trained a ResNet50 classifier using the ImageNet-1K pretrained weights to see if we can further improve the OSR performance. No moths are present in ImageNet-1K, so using the pre-trained weights should not affect evaluation on local or non-local  open-sets. Additionally, we finetuned two ViT-B-16 models \cite{dosovitskiy2021an} - one trained with ImageNet-1K and the other is the visual encoder of BioCLIP \cite{stevens2024bioclip}, a foundation model for species recognition which is increasingly used for biodiversity-related tasks. We evaluated the performance of the two models on a subset of the open-set, excluding moths that are  in its training set. Details of this subset are presented in the Appendix. 

\section{Results}

\begin{table*}[]

\caption{\textbf{Benchmarking results on Open-Insect.} We evaluate approaches falling into three categories: i) post-hoc methods, ii) training-time regularization, and iii) training with auxiliary data.
Results are shown for the three regions in Open-Insect: NE-America, W-Europe, C-America.  For each of the three open-set splits -- local (L), non-local (NL), and non-moth (NM) -- the AUROC is reported along with the accuracy of the closed-set. The best result within each category is \textbf{bold}, and the overall best result is \underline{\textbf{bold and underlined}}. For post-hoc methods, we report the mean\textsubscript{(standard deviation)} from three training runs.}
\label{tab: main result}
\begin{center}
\begin{small}
\resizebox{0.95\textwidth}{!}{%
\begin{tabular}{l|lll!{\color{gray}\vrule}l|lll!{\color{gray}\vrule}l|lll!{\color{gray}\vrule}l }
\toprule
 & \multicolumn{4}{c|}{NE-America} &   \multicolumn{4}{c|}{W-Europe} &  \multicolumn{4}{c}{C-America} \\ 
   & L & NL & NM &  Acc. & L & NL & NM &  Acc. & L & NL & NM&  Acc.  \\
  \midrule
  \multicolumn{13}{l}{\textbf{Post-hoc methods} }\\
\rowcolor{Moccasin}OpenMax \cite{bendale2016towards}&70.0 \textsubscript{(1.3)} & 75.9 \textsubscript{(1.8)} & 63.8 \textsubscript{(3.8)} & & 68.9 \textsubscript{(0.4)} & 75.9 \textsubscript{(0.8)} & 60.0 \textsubscript{(1.6)} & & 81.9 \textsubscript{(0.3)} & 82.7 \textsubscript{(0.3)} & 86.6 \textsubscript{(0.3)} & \\
\rowcolor{Moccasin}MSP \cite{hendrycks2016baseline}&86.7 \textsubscript{(0.3)} & 93.6 \textsubscript{(0.2)} & 94.3 \textsubscript{(0.3)} & & \underline{\textbf{86.0}} \textsubscript{(0.7)} & 93.2 \textsubscript{(0.6)} & 94.5 \textsubscript{(0.0)} & & 86.7 \textsubscript{(0.2)} & 88.0 \textsubscript{(0.2)} & 89.0 \textsubscript{(0.4)} & \\
\rowcolor{Moccasin}TempScale \cite{guo2017calibration}& \textbf{86.8} \textsubscript{(0.3)} & \textbf{93.7} \textsubscript{(0.2)} & 94.1 \textsubscript{(0.3)} & & \underline{\textbf{86.0}}\textsubscript{(0.6)} & \textbf{93.5} \textsubscript{(0.5)} & \textbf{94.5} \textsubscript{(0.1)} & & 85.0  \textsubscript{(1.76)} & 86.5  \textsubscript{(1.45)} & 89.3 \textsubscript{(0.62)} & \\
\rowcolor{Moccasin}ODIN \cite{liang2017enhancing}&84.7 \textsubscript{(0.3)} & 91.5 \textsubscript{(0.6)} & 89.2 \textsubscript{(0.5)} & & 81.4 \textsubscript{(3.2)} & 89.5 \textsubscript{(2.7)} & 86.1 \textsubscript{(3.6)} & & 83.5 \textsubscript{(0.3)} & 84.8 \textsubscript{(0.3)} & 85.6 \textsubscript{(0.3)} & \\
\rowcolor{Moccasin}MDS \cite{lee2018simple}&73.6 \textsubscript{(1.4)} & 79.2 \textsubscript{(1.7)} & \textbf{94.4} \textsubscript{(0.7)} & & 71.7 \textsubscript{(1.4)} & 77.1 \textsubscript{(2.0)} & 94.1 \textsubscript{(1.4)} & & 82.1 \textsubscript{(0.3)} & 83.5 \textsubscript{(0.4)} & 92.1 \textsubscript{(0.4)} & \\
\rowcolor{Moccasin}MDSEns \cite{lee2018simple}&53.2 \textsubscript{(0.2)} & 53.6 \textsubscript{(0.2)} & 62.2 \textsubscript{(0.7)} & & 50.5 \textsubscript{(0.0)} & 56.0 \textsubscript{(0.2)} & 62.5 \textsubscript{(0.4)} & & 58.2 \textsubscript{(0.1)} & 60.7 \textsubscript{(0.4)} & 69.3 \textsubscript{(0.6)} & \\
\rowcolor{Moccasin}RMDS \cite{ren2021simple}&79.9 \textsubscript{(1.5)} & 87.9 \textsubscript{(1.3)} & 90.4 \textsubscript{(3.2)} & & 79.2 \textsubscript{(1.9)} & 88.2 \textsubscript{(1.4)} & 90.5 \textsubscript{(4.9)} & & 85.0 \textsubscript{(0.3)} & 87.1 \textsubscript{(0.2)} & 91.9 \textsubscript{(0.4)} & \\
\rowcolor{Moccasin}Gram \cite{pmlr-v119-sastry20a}& - & - & - & &  - & - & - & & 46.1 \textsubscript{(0.0)} & 50.9 \textsubscript{(0.0)} & 50.5 \textsubscript{(0.0)} & \\
\rowcolor{Moccasin}EBO \cite{liu2020energy}&72.8 \textsubscript{(3.0)} & 78.2 \textsubscript{(4.0)} & 91.0 \textsubscript{(3.6)} & & 69.7 \textsubscript{(3.2)} & 76.0 \textsubscript{(3.4)} & 88.8 \textsubscript{(6.7)} & & 85.3 \textsubscript{(0.3)} & 86.7 \textsubscript{(0.4)} & 91.8 \textsubscript{(0.3)} & \\
\rowcolor{Moccasin}GradNorm \cite{huang2021importance}&49.3 \textsubscript{(2.5)} & 52.0 \textsubscript{(3.3)} & 22.9 \textsubscript{(2.7)} & & 51.1 \textsubscript{(0.5)} & 52.5 \textsubscript{(1.2)} & 25.1 \textsubscript{(8.2)} & & 33.0 \textsubscript{(1.0)} & 32.1 \textsubscript{(1.0)} & 18.3 \textsubscript{(0.5)} & \\
\rowcolor{Moccasin}ReAct \cite{sun2021react}&73.1 \textsubscript{(2.9)} & 79.9 \textsubscript{(3.8)} & 91.1 \textsubscript{(2.1)} & & 70.0 \textsubscript{(2.6)} & 78.7 \textsubscript{(2.6)} & 89.6 \textsubscript{(5.2)} & & 83.6 \textsubscript{(0.4)} & 85.1 \textsubscript{(0.5)} & 91.4 \textsubscript{(0.4)} & \\
\rowcolor{Moccasin}MLS \cite{hendrycks2019scaling}&73.8 \textsubscript{(3.1)} & 79.4 \textsubscript{(4.1)} & 91.4 \textsubscript{(3.5)} & & 70.7 \textsubscript{(3.4)} & 77.2 \textsubscript{(3.6)} & 89.3 \textsubscript{(6.5)} & & 86.5 \textsubscript{(0.3)} & 87.9 \textsubscript{(0.3)} & 91.6 \textsubscript{(0.4)} & \\
\rowcolor{Moccasin}KLM \cite{pmlr-v162-hendrycks22a}& - & - & - & &  - & - & - & & 83.5 \textsubscript{(0.3)} & 84.7 \textsubscript{(0.4)} & 88.3 \textsubscript{(0.3)} & \\
\rowcolor{Moccasin}VIM \cite{wang2022vim}&72.5 \textsubscript{(4.5)} & 78.0 \textsubscript{(5.7)} & 83.4 \textsubscript{(10.2)} & & 68.6 \textsubscript{(5.3)} & 76.3 \textsubscript{(5.1)} & 80.5 \textsubscript{(14.8)} & & 83.3 \textsubscript{(0.3)} & 84.8 \textsubscript{(0.3)} & \underline{\textbf{92.5}}\textsubscript{(0.4)} & \\
\rowcolor{Moccasin}KNN \cite{sun2022out}&64.3 \textsubscript{(17.6)} & 69.0 \textsubscript{(21.7)} & 70.3 \textsubscript{(35.3)} & & 67.3 \textsubscript{(11.4)} & 71.1 \textsubscript{(19.6)} & 74.8 \textsubscript{(28.0)} & & 80.9 \textsubscript{(0.2)} & 81.8 \textsubscript{(0.3)} & 88.3 \textsubscript{(0.3)} & \\
\rowcolor{Moccasin}DICE \cite{sun2022dice}&59.2 \textsubscript{(11.6)} & 62.9 \textsubscript{(15.4)} & 74.6 \textsubscript{(26.4)} & & 57.3 \textsubscript{(10.9)} & 62.3 \textsubscript{(15.3)} & 78.4 \textsubscript{(21.3)} & & 23.2 \textsubscript{(0.5)} & 21.5 \textsubscript{(0.6)} & 13.8 \textsubscript{(0.7)} & \\
\rowcolor{Moccasin}RankFeat \cite{song2022rankfeat}&61.3 \textsubscript{(3.6)} & 64.8 \textsubscript{(4.8)} & 69.7 \textsubscript{(13.7)} & & 52.6 \textsubscript{(2.4)} & 51.3 \textsubscript{(6.5)} & 63.5 \textsubscript{(8.2)} & & 72.3 \textsubscript{(1.1)} & 73.3 \textsubscript{(1.5)} & 86.4 \textsubscript{(1.2)} & \\
\rowcolor{Moccasin}ASH \cite{djurisic2022extremely}&72.5 \textsubscript{(2.9)} & 78.0 \textsubscript{(3.7)} & 91.5 \textsubscript{(3.4)} & & 69.2 \textsubscript{(3.2)} & 75.9 \textsubscript{(2.9)} & 89.1 \textsubscript{(6.6)} & & 82.9 \textsubscript{(0.5)} & 84.5 \textsubscript{(0.5)} & 92.0 \textsubscript{(0.3)} & \\
\rowcolor{Moccasin}SHE \cite{zhang2022out}&58.5 \textsubscript{(14.4)} & 62.9 \textsubscript{(18.4)} & 67.5 \textsubscript{(30.2)} & & 58.4 \textsubscript{(9.9)} & 61.2 \textsubscript{(15.6)} & 68.5 \textsubscript{(24.3)} & & 78.6 \textsubscript{(0.4)} & 79.4 \textsubscript{(0.8)} & 87.7 \textsubscript{(0.2)} & \\
\rowcolor{Moccasin}NECO \cite{ammar2023neco}&51.2 \textsubscript{(2.6)} & 55.2 \textsubscript{(4.3)} & 52.2 \textsubscript{(3.7)} & &  - & - & - & & 69.7 \textsubscript{(0.3)} & 70.0 \textsubscript{(0.2)} & 66.8 \textsubscript{(0.6)} & \\
\rowcolor{Moccasin}FDBD \cite{liu2023fast}&80.5 \textsubscript{(3.2)} & 88.1 \textsubscript{(3.9)} & 93.8 \textsubscript{(2.2)} & & 78.2 \textsubscript{(5.1)} & 86.1 \textsubscript{(4.9)} & 91.3 \textsubscript{(5.0)} & & 86.0 \textsubscript{(0.3)} & 87.8 \textsubscript{(0.2)} & 91.6 \textsubscript{(0.4)} & \\
\rowcolor{Moccasin}RP\textsubscript{MSP} \cite{jiang2023detecting}&86.7 \textsubscript{(0.3)} & 93.6 \textsubscript{(0.2)} & 94.3 \textsubscript{(0.3)} & & \underline{\textbf{86.0}} \textsubscript{(0.7)} & 93.2 \textsubscript{(0.6)} & \textbf{94.5} \textsubscript{(0.0)} & & \textbf{86.8} \textsubscript{(0.2)} & \textbf{88.1} \textsubscript{(0.2)} & 89.1 \textsubscript{(0.4)} & \\
\rowcolor{Moccasin}RP\textsubscript{ODIN}  \cite{jiang2023detecting}&81.7 \textsubscript{(1.1)} & 91.5 \textsubscript{(0.8)} & 90.7 \textsubscript{(0.9)} & & 82.1 \textsubscript{(1.4)} & 91.4 \textsubscript{(1.3)} & 92.3 \textsubscript{(1.5)} & & 84.2 \textsubscript{(0.2)} & 86.5 \textsubscript{(0.2)} & 89.1 \textsubscript{(0.2)} & \\
\rowcolor{Moccasin}RP\textsubscript{EBO}  \cite{jiang2023detecting}&70.8 \textsubscript{(3.6)} & 74.6 \textsubscript{(4.9)} & 89.3 \textsubscript{(4.5)} & & 67.7 \textsubscript{(3.4)} & 72.5 \textsubscript{(3.8)} & 86.7 \textsubscript{(8.0)} & & 85.7 \textsubscript{(0.3)} & 87.6 \textsubscript{(0.4)} & \underline{\textbf{92.5}} \textsubscript{(0.3)} & \\
\rowcolor{Moccasin}RP\textsubscript{GradNorm}  \cite{jiang2023detecting}&39.6 \textsubscript{(2.9)} & 38.4 \textsubscript{(4.1)} & 17.9 \textsubscript{(4.3)} & & 43.6 \textsubscript{(1.3)} & 41.4 \textsubscript{(1.0)} & 21.4 \textsubscript{(8.2)} & & 21.4 \textsubscript{(0.5)} & 20.0 \textsubscript{(0.5)} & 12.7 \textsubscript{(0.4)} & \\
\rowcolor{Moccasin}NCI \cite{liu2023detecting}&72.1 \textsubscript{(5.3)} & 78.4 \textsubscript{(5.5)} & 91.6 \textsubscript{(3.4)} &\multirow{ -26}{*}{$\underset{(0.1)}{89.7}$}& 70.9 \textsubscript{(2.9)} & 77.4 \textsubscript{(3.2)} & 89.8 \textsubscript{(6.0)} &\multirow{ -26}{*}{$\underset{(0.4)}{88.4 }$}& 85.1 \textsubscript{(0.3)} & 87.0 \textsubscript{(0.3)} & 92.4 \textsubscript{(0.4)} &\multirow{ -26}{*}{$\underset{(0.4)}{85.4}$}\\
\midrule
\multicolumn{13}{l}{\textbf{Training regularization} }\\
\rowcolor{LemonChiffon1} ConfBranch \cite{devries2018learning} & 67.2 & 70.5 & 93.8 & 89.8 & 61.4 & 59.2 & 90.8 & 86.0 & 79.8 & 81.9 & 91.0 & 85.5 \\
\rowcolor{LemonChiffon1} OpenGAN \cite{kong2021opengan} & 41.1 & 41.7 & 12.7 &88.6 & 52.7 & 56.9 & 28.2 &87.9 & 46.1 & 48.6 & 40.6 & 82.2 \\
\rowcolor{LemonChiffon1} LogitNorm \cite{wei2022mitigating} & 80.6 & 87.3 & 95.3 & 85.5 & 80.8 & 87.7 & 95.6 & 84.5 & \textbf{87.6} & \textbf{89.5} & 90.5 & \textbf{85.7} \\
\rowcolor{LemonChiffon1} ARPL \cite{chen2021adversarial} & \textbf{82.1} & 87.5 & 93.6 & 89.7 & \textbf{81.4} & 86.2 & 93.2 & 88.6 & 85.8 & 87.3 & 89.8 & 85.1 \\
\rowcolor{LemonChiffon1} G-ODIN \cite{hsu2020generalized} & 80.7 & \textbf{88.7} & 93.6 & \textbf{90.0} & 80.6 & \textbf{89.2} & \underline{\textbf{96.5}} & \textbf{88.7} & 72.7 & 74.4 & 71.4 & 83.1 \\
\rowcolor{LemonChiffon1} RotPred \cite{hendrycks2019using} & 77.6 & 85.7 & \underline{\textbf{95.9}} & 89.7 & 73.9 & 85.9 & 96.2 & 88.6 & 82.2 & 84.9 & \textbf{91.3} & 85.3 \\
\midrule
\multicolumn{13}{l}{\textbf{Training with auxiliary data } }\\
\rowcolor{AliceBlue} OE \cite{hendrycks2018deep} & 79.8 & 86.3 & 90.1 & 85.3 & 75.4 & 84.6 & 89.6 & 83.6 & 89.5 & \underline{\textbf{94.0}} & \textbf{92.1} & 84.0 \\
\rowcolor{AliceBlue} UDG \cite{yang2021semantically} & 75.1 & 81.9 & 91.5 & 80.8 & - & - & - & - & 80.8 & 83.1 & 87.9 & 76.0  \\
\rowcolor{AliceBlue} MixOE \cite{zhang2023mixture} & 86.2 & 92.5 & \textbf{94.4} & \underline{\textbf{90.3}} & \textbf{85.2} & 91.9 & \textbf{94.1} & \underline{\textbf{89.1}} & 86.2 & 87.9 & 90.1 & 84.9  \\
\rowcolor{AliceBlue} Energy \cite{liu2020energy} & \underline{\textbf{87.4}} & \underline{\textbf{95.1}} & 92.5 & 89.6 & 84.6 & \underline{\textbf{94.8}} & 89.5 & 88.4 & \underline{\textbf{90.0}} & 93.8 & 91.2 & 84.2  \\
\rowcolor{AliceBlue} NovelBranch (Ours) & 85.5 & 94.1 & 90.0 & 89.9 & 83.9 & 93.8 & 91.7 & 88.6 & 87.8 & 89.7 & 91.1 & 85.6 \\
\rowcolor{AliceBlue} Extended (Ours) & 83.5 & 92.2 & 86.1 & 89.8 & 82.6 & 91.9 & 89.4 & 88.6 & 86.9 & 89.0 & 89.3 & \underline{\textbf{85.8}}  \\
\bottomrule
\end{tabular}
}
\end{small}
\end{center}
\vskip -1em
\end{table*}

We present the performance of 38 OSR methods in Table~\ref{tab: main result}. Following OpenOOD v1.5 \cite{zhang2023openood}, we consider open-set samples as positive and closed-set samples as negative, and use the area under the ROC curve (AUROC) as the evaluation metric for OSR. We report the classification accuracy (Acc.) for the closed-set and AUROC for OSR for each method on the Open-Insect benchmark. Some of the methods listed in Table \ref{tab: main result} under ``Training regularization'' and ``Training with auxiliary data'' require combination with a post-hoc method of the user's choice. In these cases, we tested 4 simple yet effective post-hoc methods:  MSP\cite{hendrycks2016baseline}, EBO \cite{liu2020energy}, MLS \cite{hendrycks2019scaling}, and TempScale \cite{guo2017calibration}. Results in Table \ref{tab: main result} show the best performing method of each. Full results are listed in the Appendix. We also report more recently proposed OSR metrics, including OSCR \cite{dhamija2018reducing}, AUOSCR \cite{vaze2021open}, OpenAUC \cite{wang2022openauc}, and OOSA \cite{cruz2024operational} in the Appendix. Some results are shown as ``-'' either due to extremely long inference time (KLM) or very large memory requirement (Gram, NECO and UDG).

\textbf{Difficulty of the three open-set types.} Fig. \ref{fig:teaser_boxplot_examples} summarizes results for each of the three types of open-sets in each of the three regions. The maxima shown in the box plots show a clear trend that local open-set species are substantially more difficult to detect than non-local and non-moth open-set species. This validates our hypothesis that novel species from the same geographical region are more challenging to detect. One explanation for this is that open-set species that are geographically closer are also, on average, taxonomically closer to the species seen during training (see Table~\ref{tab:geographic_taxonomic_distances}). Taxonomically closer open-set species are harder to detect as they tend to exhibit more visual similarity with closed-set species \cite{vaze2021open,lang2024coarse}. However, for all regions, the best performing method can achieve over 85\% AUROC, indicating strong potential for automating species discovery with OSR in computer vision.

\textbf{Effect of pre-trained weights.} 
In general, we observe increased AUROC and closed-set accuracy when we utilize the ImageNet-1K pre-trained weights to initialize the weights of the base classifier.  All of them outperformed their training-from-scratch counterpart. Our results show that for small-scale dataset, leveraging pre-trained weights can be helpful for OSR, even if neither the closed-set nor the open-set categories are included in the pre-training dataset. Our proposed methods achieve the two highest AUROC on local open-set, 91.19 (Extended) and 92.05 (NovelBranch) and LogitNorm has the best performance of non-local open-set, with the AUROC increasing from 94.04 to 95.38 (see the Appendix for the full result). When comparing the BioCLIP weights with the ImageNet-1K weights of ViT-B-16, we find that BioCLIP weights give better performance on a subset of Open-Insect, where open-set species present in BioCLIP’s training set are excluded (see Table~\ref{tab:finetuned_vit_result}).

\begin{wraptable}{r}{0.5\linewidth}
\caption{\textbf{Comparison of pretrained weights.} OSR performance of models finetuned from BioCLIP or ImageNet-1K.  }
\label{tab:finetuned_vit_result}
\begin{center}
\resizebox{0.5\textwidth}{!}{%
\begin{tabular}{ccrrr}
\toprule
Pre-trained Weights & Method	& L & NL &  Acc.  \\
\midrule
\multirow{3}{*}{BioCLIP} & MSP &90.22	&91.61&\multirow{3}{*}{91.14}\\
&MLS &90.7	&92.7\\
&EBO&90.58	&92.6 \\
\midrule
\multirow{3}{*}{ImageNet-1K} &MSP&85.02	&85.39&\multirow{3}{*}{83.59} \\
&MLS&85.59&	86.31 \\
&EBO&85.2&	85.98 \\
\bottomrule
\end{tabular}
}
\end{center}
\vskip -0.1in
\end{wraptable}

\subsection{Post-hoc Methods}
Though no method consistently outperforms the others across datasets, we observe that maximum softmax probability (MSP) \cite{hendrycks2016baseline} and MSP-based methods, namely TempScale \cite{guo2017calibration} and RP\textsubscript{MSP} \cite{jiang2023detecting}, remain a strong baseline. For open-set moth species (both local and non-local), TempScale performs the best for NE-America and W-Europe and RP\textsubscript{MSP} performs the best for C-America. 

\textbf{Stability.} Most post-hoc methods show stable performance across the three training runs, with standard deviations below 5. However, KNN \cite{sun2022out}, DICE \cite{sun2022dice}, and SHE \cite{zhang2022out} have much higher standard deviation, as for some runs, the AUROC is smaller than 50, indicating that the OSR scores of positive (open-set) samples are generally lower than negative (closed-set) samples. We present the complete results of the three runs of these methods in the Appendix.

\textbf{Inference efficiency.} For species discovery, OSR methods are expected to be applied to a continuous stream of data over the long term, making inference efficiency a key consideration. Besides, the domain of biodiversity monitoring is often constrained by limited computing resources.  Therefore, we also evaluate the efficiency of the post-hoc methods at inference time. OpenMax \cite{bendale2016towards}, MDS \cite{lee2018simple}, and RMDS \cite{ren2021simple} are slower than other methods, and KLM \cite{pmlr-v162-hendrycks22a} is notably much slower. We list the set-up time and inference time of all post-hoc methods in the Appendix. Setup only needs to be performed once during deployment, while inference must continue to run as new data arrive. The table also indicates which methods require access to training or validation data during setup. This information can be important for end-users, as these data may not always be available, and the requirements to process them can be complicated.

\subsection{Training-time Regularization} 
\looseness=-1 These methods require further training, but they only utilize the closed-set during training time. For regions with more closed-set data (NE-America and W-Europe), training-time regularization does not improve the performance of local or non-local OSR, but it can slightly improve the performance of non-moth OSR. The opposite holds for regions with a much smaller closed-set (C-America). 

Recent post-hoc and training-time regularization methods have been shown \cite{yang2022openood} to have a substantial improvement over the MSP baseline for coarser-grained OOD detection, such as using ImageNet \cite{russakovsky2015imagenet} for ID and iNaturalist \cite{van2018inaturalist} for OOD. 
The inconsistent performance observed with the Open-Insect benchmark highlights the value of evaluating methods with a fine-grained biodiversity dataset focused on semantic shift rather than inferring performance only from more generic OOD datasets.

\subsection{Auxiliary Data}

\textbf{Realistic selection of auxiliary data helps.} Methods using auxiliary data exhibit an increase in performance for local and non-local OSR in Table \ref{tab: main result}, especially for C-America, on which there is an increase of 3.28 percentage points (pp). We believe the efficacy of the auxiliary data is due to its realistic nature. For biodiversity monitoring, expert knowledge can be used to choose species that might be similar to open-set species -- for example, moths from other regions. 

To empirically verify this claim, we train the Energy method for C-America while replacing our auxiliary data with TinyImageNet (TIN) \cite{torralba200880}, and compare the OSR performance in Table \ref{tab: result_with_tiny_image_net}. We chose TIN as it is the auxiliary data used by OpenOOD \cite{yang2022openood} for all 4 open-sets. We find that using TIN reduces OSR performance significantly, by 4.82 pp on local (L) and 5.54 pp on non-local (NL) open-sets. This suggests the importance of realistic auxiliary data selection for training. 

\begin{wraptable}{r}{0.45\linewidth}
\caption{\textbf{Ablation study on auxiliary datasets.} We report AUROC for OSR performance and accuracy on the closed-set. AUX is the auxiliary dataset used for training. TIN = TinyImageNet, OI-CA = Open-Insect C-America.}
\label{tab: result_with_tiny_image_net}
\begin{center}
\resizebox{0.4\textwidth}{!}{%
\begin{tabular}{ccccc}
\toprule
AUX & L & NL & NM & Acc. \\
\midrule
TIN	& 85.25	& 88.16& 93.85&	84.17\\
OI-CA &	90.07 &	93.7&91.33&84.19\\
\bottomrule
\end{tabular}
}
\end{center}
\end{wraptable}

\textbf{Species diversity matters in auxiliary data.} We also investigate the impact of species diversity on OSR performance. We fix the total number of images and vary the number of species and images per species. We report results for C-America using the proposed NovelBranch approach with pretrained ImageNet weights (which performs best for the local open-set). We observe that the OSR performance improves as the number of species in the auxiliary dataset increases for local novel species, but has slightly less effect for non-local open-set species (see the Appendix). This suggests that data curation efforts may focus on increasing the number of species.

\begin{wrapfigure}{r}{0.45\textwidth}
\caption{\textbf{TPR@5 (BCI) vs. AUROC (C-America O-L post-hoc methods).} Overall, models that perform well on C-America O-L also tend to achieve higher performance on the BCI data.}
\centerline{\includegraphics[width=0.4\textwidth]{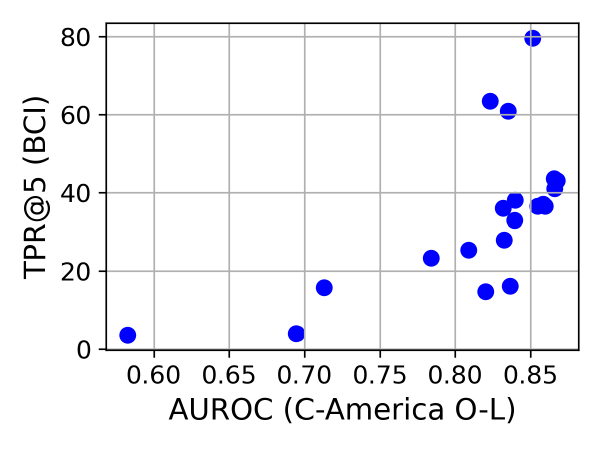}}
\label{fig: bci result}
\end{wrapfigure}

\subsection{Performance on BCI Data}

To test if the performance on the Open-Insect Benchmark can be generalized to likely undescribed species in the wild, we evaluate the post-hoc methods with the BCI images, a highly realistic test case for species discovery. While AUROC captures the overall performance of a model across all thresholds, a specific threshold must be selected to determine whether a species is from the open-set in real-world scenarios. 

Given the limited number of taxonomists available to verify new species and the large scale of data collected by camera traps during long-term monitoring, it is crucial to select a threshold that ensures a reasonable False Positive Rate (FPR) while maintaining a high True Positive Rate (TPR), or recall. Here, we consider open-set species as positive and closed-set species as negative. To achieve this goal, we consider the following metrics: the TPR of detecting O-BCI images when the supremum of the FPR of flagging closed-set species as open-set is $\tau$, denoted by TPR@$\tau$.  We visualize the result in Fig.~\ref{fig: bci result}: the x-axis gives the AUROC of different methods when evaluated on the C-America local open-set, while the y-axis gives the TPR (rate of correctly detecting open-set species), where the threshold is set so that the FPR (misclassifications of closed-set species as open) is lower than 5\%. We observe that post-hoc models that perform well on C-America O-L (high AUROC) also tend to achieve higher performance (high TPR@5) on BCI, indicating the OSR performance on Open-Insect is a good approximation for species discovery in the wild.

\subsection{Explainability of the OSR Methods}
\begin{wraptable}{r}{0.45\linewidth}
\vskip -2.0em
\caption{\textbf{OSR performance of MPS on different test data types.} Performance dropped drastically when the insect was masked in the image.}
\label{tab:explainability}
\begin{center}
\resizebox{0.4\textwidth}{!}{%
\begin{tabular}{ccccc}
\toprule
Test Data Type	& AUROC  & 	Acc. \\
\midrule
Original image &	80.01	& 71.01\\
Moth masked	& 46.57 &	1.63\\
Background masked	&76.01&	64.14\\
\bottomrule
\end{tabular}
}
\end{center}
\end{wraptable}

Explainability of the OSR methods is very important when encouraging domain experts to adopt ML-based tools for species discovery. One common concern is that models only use background features instead of species-level fine-grained features to determine whether the species from the open-set or not. We conducted an experiment to empirically verify that \textit{background features are not enough to achieve good OSR performance on Open-Insect}. Specifically, we masked the insect or the background in a subset of images (see the Appendix for details). 

We ran our C-America classifier on this subset and evaluated the  closed-set classification accuracy and the OSR performance of MSP, a well-performing post-hoc method. We observe that using the original images gives the highest closed-set accuracy (Acc.) and the best OSR performance (AUROC) (Table \ref{tab:explainability}). When the insect was masked, the model was forced to rely solely on the background and the insect's silhouette. Acc.\ dropped to just 1.63\%, and AUROC decreased to 46.57\%, close to 50\% (the performance of a random classifier). When the background was masked, the model still achieved performance comparable to that on the original images, though Acc. and AUROC decreased by 4\% and 6.87\%, respectively. When masking the background, parts of the insect such as legs and antennae were also inadvertently masked, which may explain the slight  drop.

\section{Conclusion}
\label{sec: conclusion}
We present the Open-Insect benchmark, a large-scale, fine-grained dataset focused on highly similar insect species. 
By minimizing covariate shifts of the auxiliary and open-sets from the closed-sets, our dataset allows for a more rigorous evaluation of OSR methods and will provide valuable insights for the future development of these methods. We evaluated 38 OSR methods and found that simple, efficient post-hoc approaches can perform well, achieving over 85\% AUROC in detecting fine-grained open-set species.  Besides, we show that selecting auxiliary data based on expert knowledge can further improve the OSR performance, with species diversity in the auxiliary dataset playing a key role in performance gains. Our findings also show that methods effective on the Open-Insect benchmark appear to generalize well to images of actual undescribed species. 

\textbf{Positive impacts.} We hope that the Open-Insect benchmark will draw attention to the problem of  species discovery and enable further work within the ML research community on OSR and OOD detection methods for biodiversity. Such work stands to benefit the biodiversity protection efforts across the world, the fight against climate change (which is exacerbated by biodiversity loss), and the preservation of ecosystem services on which humanity depends.

\textbf{Potential negative impacts and safeguards.} We emphasize that computer vision tools for biodiversity should \emph{not} be seen as a replacement for domain experts in ecology (who are already very limited in number) -- rather as a tool to enable such experts to gather and interpret data across more species, geographies, and timepoints than heretofore possible. Besides, classifiers for species recognition and OSR might be misused to identify species of high economic value, which may inadvertently facilitate illegal wildlife trafficking. It is thus important to engage with conservation experts to assess potential misuse before such models are deployed.

\textbf{Limitations.} Our benchmark only evaluates OSR performance on high-resolution images and does not include lower-resolution images, such as those from camera traps, due to the limited availability of annotated trap data. Furthermore, due to limited computational resources, we do not perform multiple training runs for ``training regularization'' and ``training with auxiliary data'' methods. 

\newpage
\section*{Acknowledgments}
This work was supported by the AI and Biodiversity Change (ABC) Global Center, which is funded by the US National Science Foundation (NSF) under Award No. 2330423 and Natural Sciences and Engineering Research Council (NSERC) of Canada under Award No. 585136. The work draws on research supported by the Social Sciences and Humanities Research Council (SSHRC). Any opinions, findings and conclusions or recommendations expressed in this material are those of the authors and do not necessarily reflect the views of the NSF, NSERC, or SSHRC.
This work was also supported in part by the Pioneer Centre for AI, DNRF grant number P1, by the Global Wetland Center (grant number NNF23OC0081089) from Novo Nordisk Foundation, by the Canada CIFAR AI Chairs program, by the Fonds de recherche du Québec – Nature et technologies (FRQNT), by the MIT-IBM Watson AI Lab Partners Program, by the MIT-Google Program for Computing Innovation, by the MIT MGAIC Seed Funds program, by NSF Award No. 2441060, and by the Schmidt Sciences AI2050 Program. 

This research was enabled in part by compute resources, software and technical help provided by Mila (mila.quebec). We thank Francis Pelletier for the valuable guidance on the code implementation and dataset preparation.


\bibliographystyle{plain}
\bibliography{refs}


\newpage
\section*{NeurIPS Paper Checklist}

\begin{enumerate}

\item {\bf Claims}
    \item[] Question: Do the main claims made in the abstract and introduction accurately reflect the paper's contributions and scope?
    \item[] Answer: \answerYes{} 
    \item[] Justification: The main claims made in the abstract and introduction summarize the paper's contributions and scope. 

\item {\bf Limitations}
    \item[] Question: Does the paper discuss the limitations of the work performed by the authors?
    \item[] Answer: \answerYes{} 
    \item[] Justification: We discussion the limitations in Sec. \ref{sec: conclusion}.

\item {\bf Theory assumptions and proofs}
    \item[] Question: For each theoretical result, does the paper provide the full set of assumptions and a complete (and correct) proof?
    \item[] Answer: \answerNA{} 
    \item[] Justification: This paper does not include theorems.

    \item {\bf Experimental result reproducibility}
    \item[] Question: Does the paper fully disclose all the information needed to reproduce the main experimental results of the paper to the extent that it affects the main claims and/or conclusions of the paper (regardless of whether the code and data are provided or not)?
    \item[] Answer:  \answerYes{} 
    \item[] Justification: We provide  training details in Sec \ref{sec: methods} as well as App.~\ref{sec: training_details}. Our code is available at \url{https://github.com/RolnickLab/Open-Insect}.  
    Our datasets are avaiable at  \url{https://huggingface.co/datasets/yuyan-chen/open-insect} and \url{https://huggingface.co/datasets/yuyan-chen/open-insect-bci}. 

\item {\bf Open access to data and code}
    \item[] Question: Does the paper provide open access to the data and code, with sufficient instructions to faithfully reproduce the main experimental results, as described in supplemental material?
    \item[] Answer: \answerYes{}{} 
    \item[] Justification: Our datasets are hosted on huggingface: \url{https://huggingface.co/datasets/yuyan-chen/open-insect} and \url{https://huggingface.co/datasets/yuyan-chen/open-insect-bci}. Our code is available at \url{https://github.com/RolnickLab/Open-Insect}.

\item {\bf Experimental setting/details}
    \item[] Question: Does the paper specify all the training and test details (e.g., data splits, hyperparameters, how they were chosen, type of optimizer, etc.) necessary to understand the results?
    \item[] Answer: \answerYes{}{} 
    \item[] Justification: We provide  training details in Sec \ref{sec: methods} as well as App.~\ref{sec: training_details}. Our code is available at \url{https://github.com/RolnickLab/Open-Insect}.

\item {\bf Experiment statistical significance}
    \item[] Question: Does the paper report error bars suitably and correctly defined or other appropriate information about the statistical significance of the experiments?
    \item[] Answer: \answerYes{}{} 
    \item[] Justification: The main results are reported in Table 2. We report the mean and standard deviation for post-hoc methods from three training runs. Other training methods were only trained once, and we discuss this limitation in Sec. \ref{sec: conclusion}.

\item {\bf Experiments compute resources}
    \item[] Question: For each experiment, does the paper provide sufficient information on the computer resources (type of compute workers, memory, time of execution) needed to reproduce the experiments?
    \item[] Answer: \answerYes{},  
    \item[] Justification: We provide these information in Sec. \ref{sec: training_details}, \ref{sec: inference requirement}, and \ref{sec: methods}.

    
\item {\bf Code of ethics}
    \item[] Question: Does the research conducted in the paper conform, in every respect, with the NeurIPS Code of Ethics \url{https://neurips.cc/public/EthicsGuidelines}?
    \item[] Answer: \answerYes{}{} 
    \item[] Justification: The research conducted in the paper conforms with the NeurIPS Code of Ethics.

\item {\bf Broader impacts}
    \item[] Question: Does the paper discuss both potential positive societal impacts and negative societal impacts of the work performed?
    \item[] Answer: \answerYes{}{} 
    \item[] Justification: We discuss both positive and negative impacts in Sec. \ref{sec: conclusion}.
    
\item {\bf Safeguards}
    \item[] Question: Does the paper describe safeguards that have been put in place for responsible release of data or models that have a high risk for misuse (e.g., pretrained language models, image generators, or scraped datasets)?
    \item[] Answer: \answerYes{} 
    \item[] Justification: We discuss safeguards in Sec. \ref{sec: conclusion}.

\item {\bf Licenses for existing assets}
    \item[] Question: Are the creators or original owners of assets (e.g., code, data, models), used in the paper, properly credited and are the license and terms of use explicitly mentioned and properly respected?
    \item[] Answer: \answerYes{} 
    \item[] Justification: Our codebase is built using OpenOOD \cite{yang2022openood}. We cite the paper of OpenOOD and include the URL of the GitHub repository in Sec. \ref{sec: methods}. Our dataset partially builds upon data from the AMI dataset \cite{jain2024insect}. We cite this paper in Sec. \ref{sec: ID data}. We include the license of the AMI dataset in the Open-Insect dataset hosting on huggingface \url{https://huggingface.co/datasets/yuyan-chen/open-insect}. 

\item {\bf New assets}
    \item[] Question: Are new assets introduced in the paper well documented and is the documentation provided alongside the assets?
    \item[] Answer: \answerYes{} 
    \item[] Justification: Our datasets are hosted on huggingface: \url{https://huggingface.co/datasets/yuyan-chen/open-insect} and \url{https://huggingface.co/datasets/yuyan-chen/open-insect-bci}. Our code is available at \url{https://github.com/RolnickLab/Open-Insect}. Details about training and limitations are discussed in the main text and the Appendix. Both datasets are distributed under the CC-BY-NC 4.0 License. The code is distributed under the MIT License. 

\item {\bf Crowdsourcing and research with human subjects}
    \item[] Question: For crowdsourcing experiments and research with human subjects, does the paper include the full text of instructions given to participants and screenshots, if applicable, as well as details about compensation (if any)? 
    \item[] Answer: \answerNA{} 
    \item[] Justification: This paper does not involve crowdsourcing nor research with human subjects. 

\item {\bf Institutional review board (IRB) approvals or equivalent for research with human subjects}
    \item[] Question: Does the paper describe potential risks incurred by study participants, whether such risks were disclosed to the subjects, and whether Institutional Review Board (IRB) approvals (or an equivalent approval/review based on the requirements of your country or institution) were obtained?
    \item[] Answer: \answerNA{} 
    \item[] Justification: This paper does not involve crowdsourcing nor research with human subjects.

\item {\bf Declaration of LLM usage}
    \item[] Question: Does the paper describe the usage of LLMs if it is an important, original, or non-standard component of the core methods in this research? Note that if the LLM is used only for writing, editing, or formatting purposes and does not impact the core methodology, scientific rigorousness, or originality of the research, declaration is not required.
    \item[] Answer: \answerNA{} 
    \item[] Justification: This research does not involve LLMs.

\end{enumerate}

\newpage
\appendix

\section*{Technical Appendices and Supplementary Material}
\section{Dataset}

\subsection{Biome list} \label{biome list}
1: ``Tropical and Subtropical Moist Broadleaf Forests",
    2: ``Tropical and Subtropical Dry Broadleaf Forests",
    3: ``Tropical and Subtropical Coniferous Forests",
    4: ``Temperate Broadleaf and Mixed Forests",
    5: ``Temperate Coniferous Forests",
    6: ``Boreal Forests/Taiga",
    7: ``Tropical and Subtropical Grasslands, Savannas, and Shrublands``,
    8: ``Temperate Grasslands, Savannas, and Shrublands",
    9: ``Flooded Grasslands and Savannas",
    10: ``Montane Grasslands and Shrublands",
    11: ``Tundra",
    12: ``Mediterranean Forests, Woodlands, and Scrub",
    13: ``Deserts and Xeric Shrublands",
    14: ``Mangroves",
    98: ``Lake",
    99: ``Rock and Ice".

\subsection{Choice of degrees in latitude and longitude}

We wanted to choose surrounding areas that are small enough so that the species can simulate undescribed species, but at the same time, large enough so that we have a relatively diverse species distribution. Since typically the variation in species is higher across latitudes than longitudes due to climatic shifts \cite{hillebrand2004generality}, we chose 1 degree in latitude and 3 degrees in longitude. We also visualize the biomes of each region in Fig.~\ref{fig: global biome map} to show that the surrounding regions are very similar to the in-distribution regions.

\subsection{Global moth data curation} \label{sec: Global moth dataset curation}
We curate the global list of moth species from GBIF~\cite{gbif_moth_checklist}. Together, the moth and butterfly families make up the order \emph{Lepidoptera}. While there are over 100 families of moths, there are only seven families of butterflies. Therefore, we exclude the butterfly families from the order Lepidoptera to get only moth families and their corresponding species. Next, we consider only those species whose taxonomic status is classified as accepted in the backbone. This process results in a total of 46,983 species. Simultaneously, we download metadata for all Lepidoptera observations that include images using GBIF’s occurrence search tool~\cite{gbif_dwca}. We then fetch images for the moth species, limiting to 1,000 images per species class. This effort yields  ~4.5 million images. Following the data cleaning procedures outlined in \cite{jain2024insect}, we remove images with duplicate URLs, problematic sources, thumbnails, and non-adult images using a life stage classifier. After these steps, we have  ~3.8 million images remaining. Additionally, we eliminate any species that have fewer than five images in the dataset and only include images with the following licenses: CC BY 4.0, CC BY-NC 4.0, CC BY-NC-SA 4.0, CC0 1.0 (Public Domain), CC BY-SA 4.0, No Rights Reserved, CC BY 3.0, CC BY-SA 3.0. This finalizes our globla data to have 28,388 species and 2,912,168 images. 

\subsection{The Open-Insect Dataset}
\begin{figure}[htbp]
  \centering
  \begin{subfigure}[b]{0.49\textwidth}
    \includegraphics[width=\textwidth]{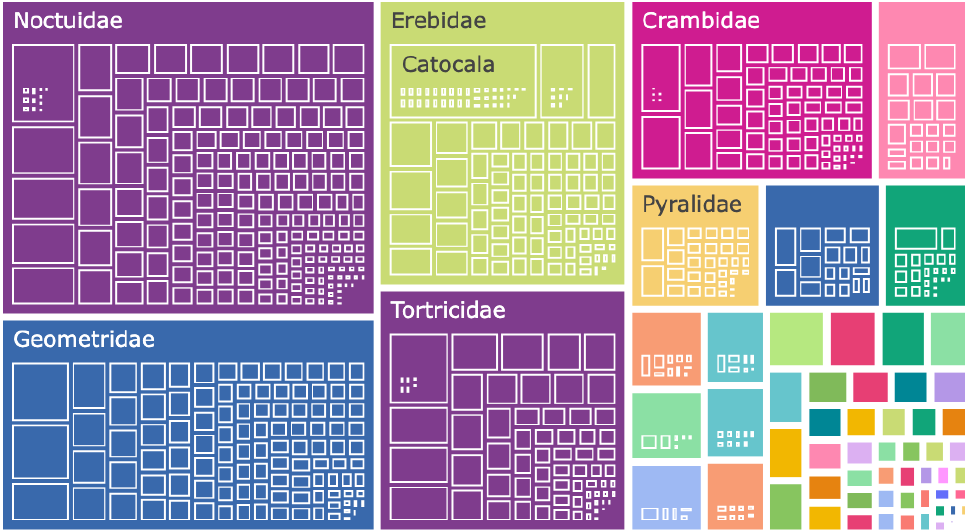}
    \caption{NE-America Closed-Set}
  \end{subfigure}
  \hfill
  \begin{subfigure}[b]{0.49\textwidth}
    \includegraphics[width=\textwidth]{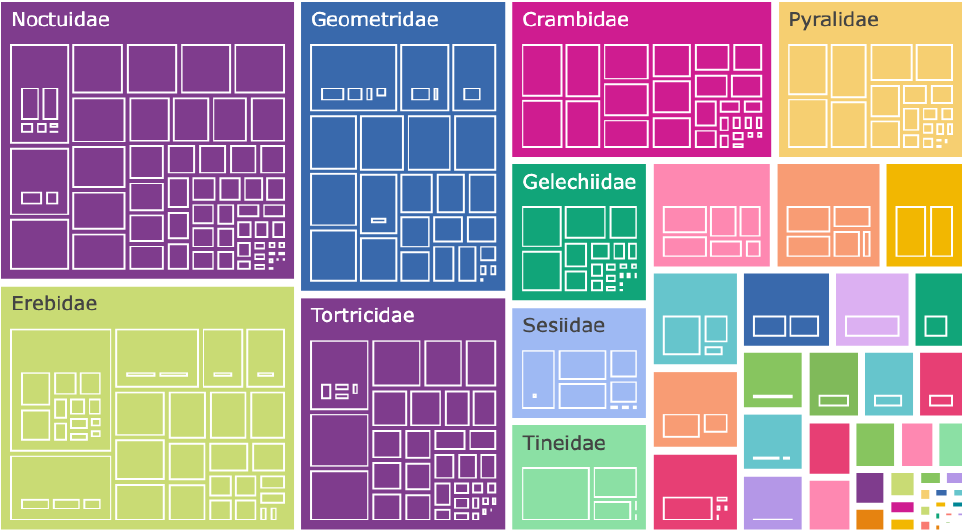}
    \caption{NE-America Open-Set Local}
  \end{subfigure}

  \vskip\baselineskip
  \begin{subfigure}[b]{0.49\textwidth}
    \includegraphics[width=\textwidth]{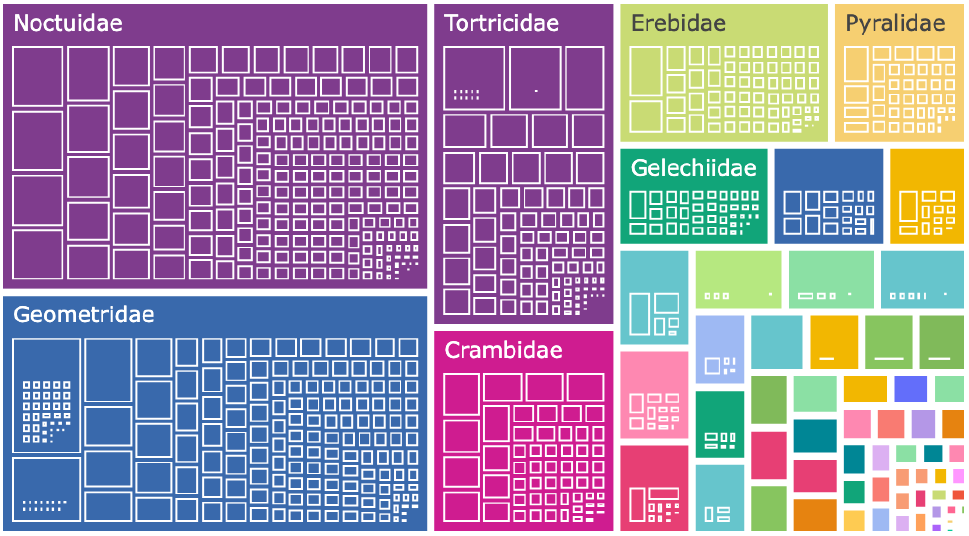}
    \caption{W-Europe Closed-Set}
  \end{subfigure}
  \hfill
  \begin{subfigure}[b]{0.49\textwidth}
    \includegraphics[width=\textwidth]{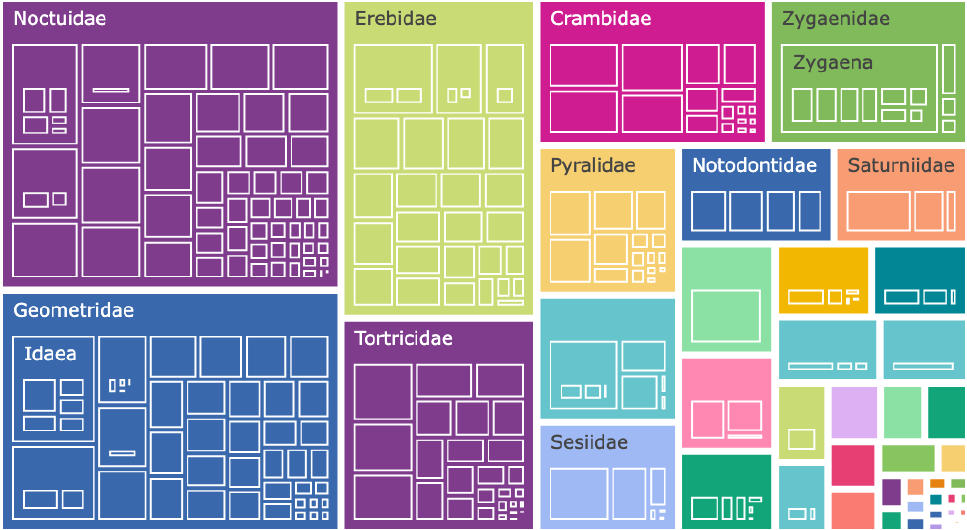}
    \caption{W-Europe Open-Set Local}
  \end{subfigure}

  \vskip\baselineskip
  \begin{subfigure}[b]{0.49\textwidth}
    \includegraphics[width=\textwidth]{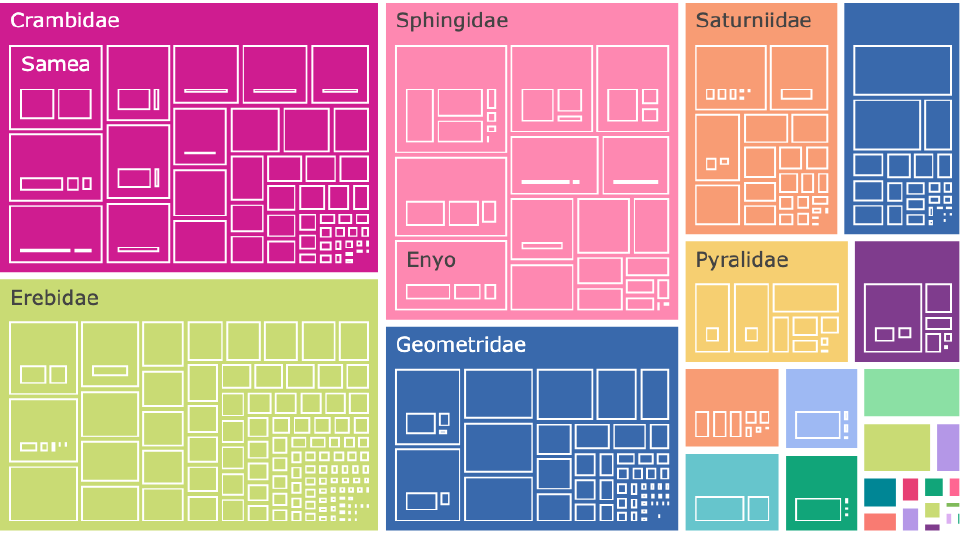}
    \caption{C-America Closed-Set}
  \end{subfigure}
  \hfill
  \begin{subfigure}[b]{0.49\textwidth}
    \includegraphics[width=\textwidth]{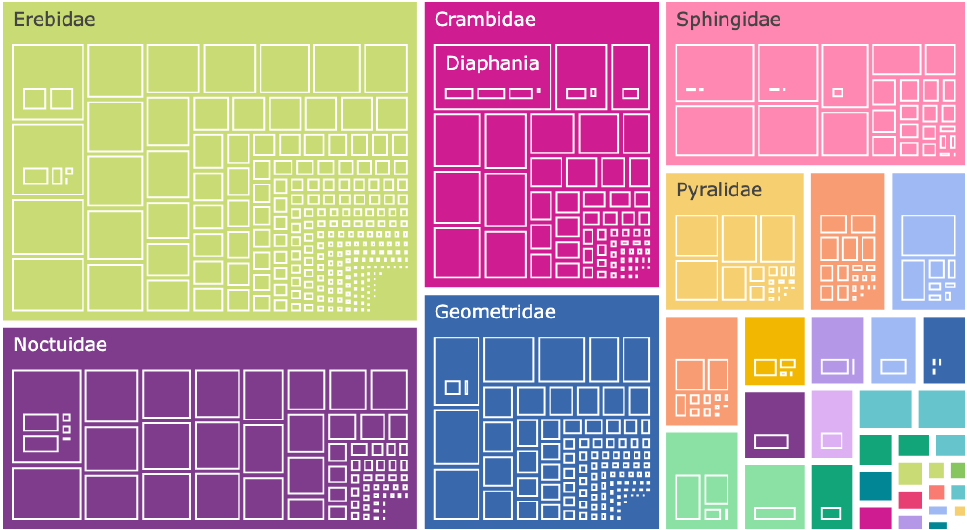}
    \caption{C-America Open-Set Local}
  \end{subfigure} 

    \vskip\baselineskip
  \begin{subfigure}[b]{0.6\textwidth}
    \includegraphics[width=\textwidth]{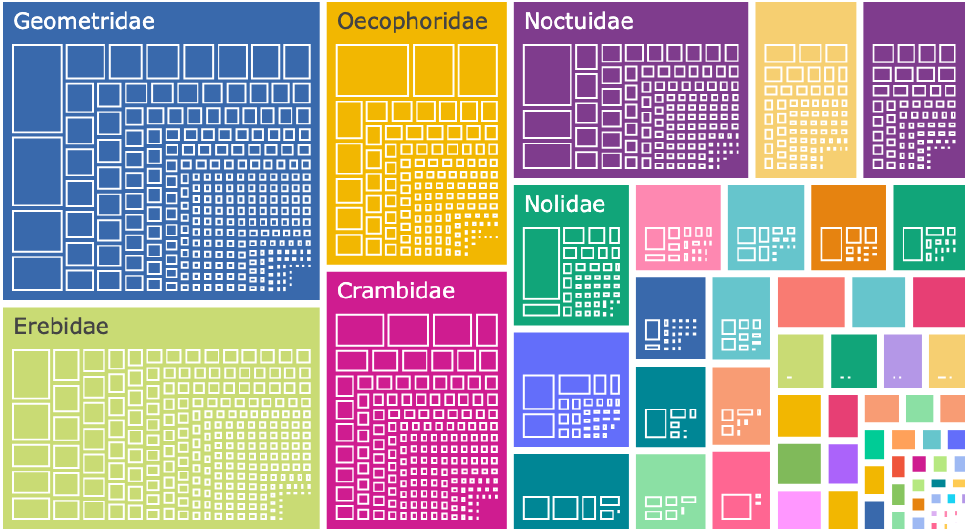}
    \caption{Non-Local Open-Set}
  \end{subfigure}

 \caption{\textbf{Visualization of the Open-Insect taxonomic distribution.} Tree maps (a)–(f) show the taxonomic composition of moth families in Open-Insect across three regions. Each nested box represents a genus or species, and box size reflects the relative number of images. The same family is \emph{colored consistently} across regions. Local open-set species display taxonomic distributions more similar to their corresponding closed-set species, indicating shared families and comparable visual traits. In contrast, the non-local open-set samples from Australia (g) exhibit markedly different taxonomic and color patterns, reflecting greater divergence from the training regions.}

  \label{fig:all_treemaps}
\end{figure}

We compare the data distribution of local and non-local open-set datasets of three regions: Northeastern North America (NE-America), Western Europe (W-Europe), and Central America (C-America) in Table \ref{tab:geographic_taxonomic_distances} and visualize the taxonomic distribution in Fig.~\ref{fig:all_treemaps}.

\begin{table*}[!htbp]
\caption{\textbf{Geographic and taxonomic distance of Open-Insect open-set splits.} Comparison of the local and non-local open-set datasets.}
\label{tab:geographic_taxonomic_distances}
\begin{center}
\resizebox{0.95\textwidth}{!}{%
\begin{tabular}{l|cccc|cccc|cccr}
\toprule
& \multicolumn{4}{c|}{NE-America} & \multicolumn{4}{c|}{W-Europe} &\multicolumn{4}{c}{C-America} \\
& \multicolumn{2}{c}{Local} & \multicolumn{2}{c|}{Non-local} & \multicolumn{2}{c}{Local} & \multicolumn{2}{c|}{Non-local} & \multicolumn{2}{c}{Local} & \multicolumn{2}{c}{Non-local} \\
& Species & Images & Species & Images & Species & Images & Species & Images & Species & Images & Species & Images\\
\midrule
1-hop  & 463 & 66,370 & 553 & 11,919 & 323 & 44,119 & 802 & 16,962 & 772 & 46,731 & 151& 3,428 \\
2-hop & 147& 46,407 & 3,810 & 77,967 & 132 & 33,796 & 3,482 & 71,355 & 908 & 61,651 & 3,238 & 68,508  \\
3-hop & 7& 857 & 299 & 5,711 & 3 & 733 & 378 & 7,280 & 44 & 3,497 & 1,273 &23,661 \\
\bottomrule
\end{tabular}
}
\end{center}
\vskip -0.1in
\end{table*}

\subsection{Comparison to related work}

\textbf{Taxonomic level.} BIOSCAN-1M (B-1M) \cite{NEURIPS2023_87dbbdc3}, BIOSCAN-5M (B-5M) \cite{gharaee2024bioscan}, and our Open-Insect, focus on insects, while Insect-1M \cite{nguyen2024insect} consist of insects and other arthropods. B-1M and B-5M mainly focus on flies and use Barcode Index Numbers (BINs) as labels (< 10\% of these BINs correspond to species, while others are higher order taxonomic groups of various levels).  Our dataset focuses on moths and all samples are identified to species level. This allows us to evaluate the difficulty of OSR in highly fine-grained recognition and at specified taxonomic levels.

\textbf{Data type.} While all datasets include images, the image types are different. Open-Insect and Insect-1M contain images of museum specimens and live insects in the wild. B-1M and B-5M consist of microscope images of specimens collected using Malaise traps. While Malaise trap specimens are often too small and mangled for species-level identification, Open-Insect was designed to support species-level identification and new species discovery, focusing on more readily identifiable camera trap images, as well as slightly larger insects that can be visually identified.

\textbf{Benchmark tasks}:  The intended applications of the other three datasets are very different from those of Open-Insect. B-1M and B-5M assess the effectiveness of multi-modality by aligning image data and DNA barcodes, while Insect-1M was curated to train a foundation model for insects and other arthropods using images and text. Open-Insect is designed for visual species identification where DNA barcodes or text description may not be available. Given the strong covariate shift and difference in tasks between B-1M / B-5M and Open-Insect, it is not possible to directly evaluate OSR performance of models trained on B-1M and B-5M using Open-Insect.

\section{Methods}

\subsection{Training details}
\label{sec: training_details}

All images were resized to 126 by 126. We followed the data pre-processing implementation in OpenOOD \cite{yang2022openood}. All models were trained with 1 RTX8000 GPU, with 16 CPUs, 16 workers, and 100 GB of CPU memory in total.

\textbf{Training from scratch.} All species recognition classifiers were trained from scratch for 120 epochs with a batch size of 512 for each region. We used the AdamW optimizer \cite{loshchilov2017decoupled} with an initial learning rate of 0.01. The learning rate was decayed with cosine scheduling \cite{he2019bag} with the first 6 epochs being warm-up epochs.

\textbf{Fine-tuning.} 
For all training methods (whether requiring extra data or not) \label{sec: finetuing}, we fine-tuned the species-level classifiers for 30 additional epochs with a batch size of 512 except for RotPred, which was trained with a batch size of 256 due to a larger memory requirement. Similarly, we used AdamW as the optimizer with cosine scheduling for learning rate scheduling. The initial learning rate was 0.001, with the first two epochs as warm-up. 

\section{Additional results}

\subsection{All metrics}
\begin{figure}[!htbp]
    \centering
    \begin{subfigure}[b]{0.32\linewidth}
        \centering
        \includegraphics[width=\linewidth]{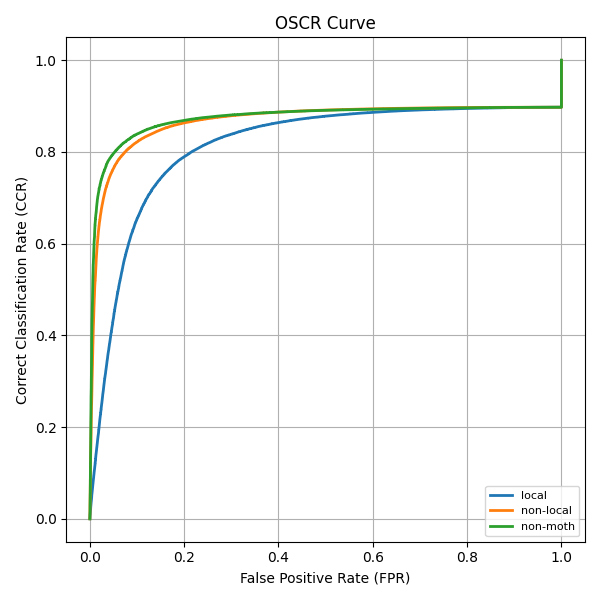}
        \caption{NE-America}
        \label{fig:oscr_ne_america}
    \end{subfigure}
    \hfill
    \begin{subfigure}[b]{0.32\linewidth}
        \centering
        \includegraphics[width=\linewidth]{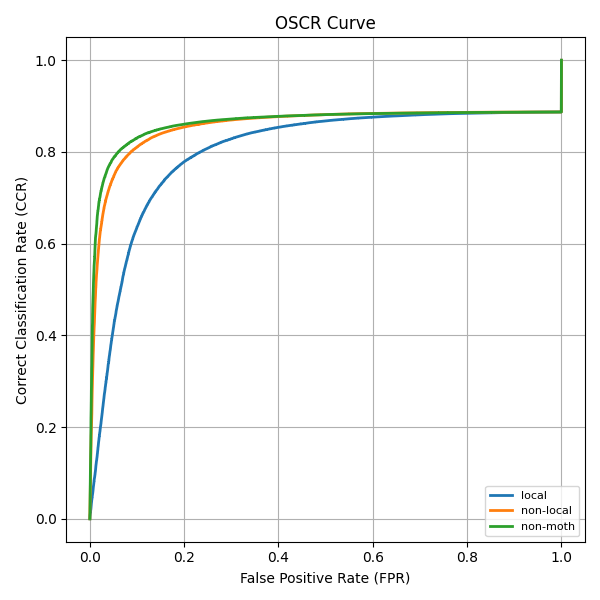}
        \caption{W-Europe}
        \label{fig:oscr_europe}
    \end{subfigure}
    \hfill
    \begin{subfigure}[b]{0.32\linewidth}
        \centering
        \includegraphics[width=\linewidth]{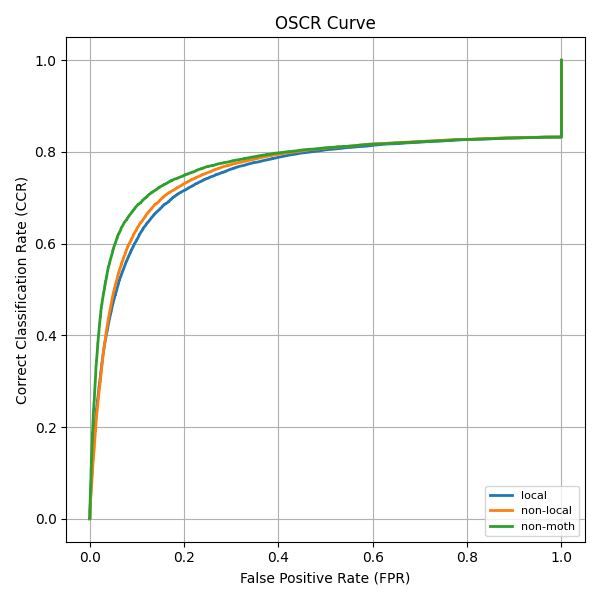}
        \caption{C-America}
        \label{fig:oscr_asia}
    \end{subfigure}
    \caption{\textbf{OSCR curves of the basic classifier with MSP as the post-hoc method across three regions.} The x-axis is the False Positive Rate (FPR) and the y-axis is the Correct Classification Rate (CCR).}
    \label{fig:oscr_regions}
\end{figure}

We show the Open-Set Classification Rate (OSCR) curve \cite{dhamija2018reducing} of the basic classifier with MSP as the post-hoc method across the three regions in Fig.~\ref{fig:oscr_regions}. We also report the Area Under the OSCR curve (AUOSCR) \cite{vaze2021open}, OpenAUC \cite{wang2022openauc}, and Operational Open-Set Accuracy (OOSA) \cite{cruz2024operational} in Tables~\ref{tab:all_metrics_ne_america}, \ref{tab:all_metrics_w_europe}, and \ref{tab:all_metrics_c_america}. AUOSCR, OpenAUC, and OOSA depend on the closed-set accuracy, whereas AUROC is independent of it. The closed-set accuracies reported in Table~\ref{tab: main result} are computed in the standard manner, using the argmax of the logit vector. Hence, all post-hoc methods have the same closed-set classification accuracy. However, some post-hoc methods also provide an alternative way to derive predictions. In such cases, we report the metrics computed using both (1) the standard approach and (2) the post-hoc prediction method (denoted with *) in the subsequent tables if the metrics are drastically different. 

\begin{table*}[!htbp]
\caption{\textbf{Full results of NE-America.} AUROC, AUOSCR, OpenAUC, and OOSA are averaged across three open-set splits: Local, Non-Local, and Non-Moth. For post-hoc methods, we present the result of one of the three runs.} 
\label{tab:all_metrics_ne_america}
\begin{center}
\begin{small}
\begin{tabular}{llllll}
\toprule
Method & AUROC & AUOSCR & OpenAUC & OOSA & ACC \\
\midrule
MSP & 91.7 & 85.1 & 85.1 & 84.5 & 89.8 \\
TempScale & 91.7 & 85.1 & 85.1 & 84.5 & 89.8 \\
ODIN & 88.9 & 81.9 & 81.9 & 81.0 & 89.7 \\
MDS & 82.2 & 77.5 & 77.5 & 77.8 & 89.8 \\

MDSEns & 55.9 &  50.7   &  50.7 & 50.0 & 89.8 \\
MDSEns* & 55.9 & 1.2 & 1.2 & 50.0 & 1.9 \\
RMDS & 87.0 & 81.2 & 81.2 & 80.6 & 89.8 \\
Gram & - & - & - & - & - \\
EBO & 82.2 & 77.4 & 77.4 & 76.8 & 89.8 \\
GradNorm & 43.8 & 37.1 & 37.1 & 50.2 & 89.8 \\
ReAct & 81.8 &   76.6 &    76.6 & 60.8 & 89.8 \\
ReAct* & 81.8 & 19.5 & 19.5 & 54.6 & 21.9 \\
MLS & 83.2 & 78.4 & 78.4 & 77.7 & 89.8 \\
KLM & - & - & - & - & - \\
VIM & 81.8 & 76.9 & 76.9 & 74.8 & 89.8 \\
KNN & 84.7 & 78.7 & 78.7 & 77.3 & 89.8 \\
DICE& 78.7 &   73.8   &  73.8 & 50.0 & 89.8 \\
DICE* & 75.0 & 7.8 & 7.8 & 51.3 & 8.8 \\
RankFeat* & 65.2 & 22.4 & 22.4 & 50.3 & 32.0 \\
RankFeat & 65.2 &   60.3  &   60.3 & 50.0 & 89.8 \\
ASH & 82.1 & 76.3 & 76.3 & 76.3 & 87.8 \\
SHE & 75.5 & 69.8 & 69.8 & 68.4 & 89.8 \\
NECO & - & - & - & - & - \\
FDBD & 89.6 & 83.9 & 83.9 & 83.2 & 89.8 \\
RP\textsubscript{MSP} & 91.7 & 85.1 & 85.1 & 84.5 & 89.8 \\
RP\textsubscript{ODIN} & 88.6 & 81.6 & 81.6 & 82.4 & 89.7 \\
RP\textsubscript{EBO} & 80.1 & 75.5 & 75.5 & 75.2 & 89.8 \\
RP\textsubscript{GradNorm} & 31.9 & 25.7 & 25.7 & 50.0 & 89.8 \\
NCI & 83.1 & 78.3 & 78.3 & 77.4 & 89.8 \\
ConfBranch & 77.1 & 72.8 & 72.8 & 72.0 & 89.8 \\
OpenGAN &31.8  &  25.8   &  25.8  &43.4 & 88.6 \\
LogitNorm & 87.7 & 79.7 & 79.7 & 81.0 & 85.5 \\
ARPL & 87.7 & 82.8 & 82.8 & 80.8 & 90.0 \\
GODIN &  87.7 &   82.0 &    82.0 &  81.7 &  90.0 \\
RotPred &  86.4 &   80.0 &    80.0 &  79.0 &  89.7 \\
OE & 85.4 & 77.4 & 77.4 & 78.8 & 85.3 \\
UDG & 82.8 & 71.4 & 71.4 & 73.7 & 80.8 \\
MixOE & 91.0 & 85.3 & 85.3 & 84.1 & 90.3 \\
Energy & 91.7 & 83.9 & 83.9 & 82.6 & 89.6 \\
NovelBranch & 89.9 & 82.5 & 82.5 & 80.4 & 89.9 \\
Extended & 87.2 & 80.2 & 80.2 & 77.7 & 89.8 \\
\bottomrule
\end{tabular}
\end{small}
\end{center}
\end{table*}

\begin{table*}[!htbp]
\caption{\textbf{Full results of W-Europe.} AUROC, AUOSCR, OpenAUC, and OOSA are averaged across three open-set splits: Local, Non-Local, and Non-Moth. For post-hoc methods, we present the result of one of the three runs.} 
\label{tab:all_metrics_w_europe}
\begin{center}
\begin{small}
\begin{tabular}{llllll}
\toprule
Method & AUROC & AUOSCR & OpenAUC & OOSA & ACC \\
\midrule
MSP & 91.5 & 84.0 & 84.0 & 83.9 & 88.7 \\
TempScale & 91.6 & 84.0 & 84.0 & 84.0 & 88.7 \\
ODIN & 88.0 & 80.2 & 80.2 & 79.5 & 88.7 \\
MDS & 81.6 & 76.0 & 76.0 & 76.6 & 88.7 \\
MDSEns & 56.3 & 50.4 & 50.4 & 51.7 & 88.7 \\
MDSEns* & 56.3 & 1.2 & 1.2 & 50.1 & 1.9 \\
RMDS & 87.7 & 80.7 & 80.7 & 80.5 & 88.7 \\
Gram & - & - & - & - & - \\
EBO & 81.3 & 75.7 & 75.7 & 75.6 & 88.7 \\
GradNorm & 41.7 & 34.5 & 34.5 & 50.0 & 88.7 \\
ReAct & 81.9 & 75.8 & 75.8 & 72.4 & 88.7 \\
ReAct* & 81.9 & 13.6 & 13.6 & 53.5 & 14.8 \\
MLS & 82.2 & 76.6 & 76.6 & 76.5 & 88.7 \\
KLM & - & - & - & - & - \\
VIM & 81.0 & 75.2 & 75.2 & 74.8 & 88.7 \\
KNN & 84.3 & 77.2 & 77.2 & 75.9 & 88.7 \\
DICE & 80.7 & 74.5 & 74.5 & 72.4 & 88.7 \\
DICE* &80.7   &  3.0   &   3.0 & 50.0 & 3.2 \\
RankFeat & 60.8  &  55.9  &   55.9 & 51.8  &88.7\\
RankFeat* & 60.8 & 30.5 & 30.5 & 53.0 & 41.5 \\
ASH & 81.3  &  75.7   &  75.7 & 72.5  &88.7\\
SHE & 73.8 & 66.9 & 66.9 & 66.4 & 88.7 \\
NECO & - & - & - & - & - \\
FDBD & 88.9 & 82.3 & 82.3 & 82.0 & 88.7 \\
RP\textsubscript{MSP} & 91.5 & 84.0 & 84.0 & 83.9 & 88.7 \\
RP\textsubscript{ODIN} & 89.7 & 81.7 & 81.7 & 83.5 & 88.6 \\
RP\textsubscript{EBO} & 79.2 & 73.9 & 73.9 & 74.1 & 88.7 \\
RP\textsubscript{GradNorm} & 33.1 & 26.4 & 26.4 & 50.0 & 88.7 \\
NCI & 82.2 & 76.5 & 76.5 & 76.2 & 88.7 \\
ConfBranch & 71.7 & 67.3 & 67.3 & 67.4 & 88.6 \\
OpenGAN & 45.9   & 40.8  &   40.8  &47.1 & 87.9\\
LogitNorm & 88.0 & 78.9 & 78.9 & 80.6 & 84.5 \\
ARPL & 86.9 & 81.1	& 81.1	& 79.2	& 89.0\\
  GODIN &  88.7 &   81.7 &    81.7 &  81.5 &  88.8 \\
RotPred & 85.3 & 77.6 & 77.6 & 77.0 & 87.9 \\
OE & 83.2 & 73.0 & 73.0 & 74.6 & 83.6 \\
UDG & - & - & - & - & - \\
MixOE & 90.4 & 83.8 & 83.8 & 82.9 & 89.1 \\
Energy & 89.7 & 81.0 & 81.0 & 79.7 & 88.4 \\
NovelBranch & 89.8 & 81.6 & 81.6 & 80.5 & 88.6 \\
Extended & 88.0 & 80.0 & 80.0 & 78.5 & 88.6 \\
\bottomrule
\end{tabular}
\end{small}
\end{center}
\end{table*}

\begin{table*}[!htbp]
\caption{\textbf{Full results of C-America.} AUROC, AUOSCR, OpenAUC, and OOSA are averaged across three open-set splits: Local, Non-Local, and Non-Moth. For post-hoc methods, we present the result of one of the three runs.} 
\label{tab:all_metrics_c_america}
\begin{center}
\begin{small}
\begin{tabular}{llllll}
\toprule
Method & AUROC & AUOSCR & OpenAUC & OOSA & ACC \\
\midrule
MSP & 87.6 & 79.5 & 79.5 & 80.6 & 85.0 \\
TempScale & 85.6 & 76.3 & 76.3 & 77.8 & 83.2 \\
ODIN & 86.4 & 78.1 & 78.1 & 42.5 & 85.0 \\
MDS & 85.8 & 77.9 & 77.9 & 79.2 & 84.6 \\
MDSEns & 62.6 & 54.5 & 54.5 & 55.8 & 85.0 \\
MDSEns* & 62.6 & 9.3 & 9.3 & 50.6 & 11.7 \\
RMDS & 85.6 & 76.2 & 76.2 & 78.6 & 83.2 \\
Gram & 50.9 & 43.1 & 43.1 & 50.0 & 85.0 \\
EBO & 87.5 & 79.6 & 79.6 & 81.4 & 85.0 \\
GradNorm & 28.9 & 21.0 & 21.0 & 50.0 & 85.0 \\
ReAct & 86.2 & 78.0 & 78.0 & 50.8 & 84.5 \\
MLS & 88.3 & 80.2 & 80.2 & 81.5 & 85.0 \\
KLM & 85.1 & 77.5 & 77.5 & 79.4 & 85.0 \\
VIM & 86.5 & 78.8 & 78.8 & 79.7 & 85.0 \\
KNN & 84.4 & 76.6 & 76.6 & 77.6 & 85.0 \\
DICE & 20.8 & 13.2 & 13.2 & 50.0 & 85.0 \\
DICE* & 20.8 & 12.1 & 12.1 & 50.0 & 82.7 \\
RankFeat & 76.6 & 69.7 & 69.7 & 69.0 & 85.0 \\
RankFeat* & 76.6 & 53.9 & 53.9 & 65.9 & 60.4 \\
ASH & 87.5 & 79.6 & 79.6 & 81.4 & 85.0 \\
SHE & 82.6 & 75.1 & 75.1 & 74.1 & 85.0 \\
NECO & 65.4 & 57.5 & 57.5 & 42.4 & 84.7 \\
FDBD & 88.1 & 80.0 & 80.0 & 81.0 & 85.0 \\
RP\textsubscript{MSP} & 87.7 & 79.6 & 79.6 & 80.6 & 85.0 \\
RP\textsubscript{ODIN} & 87.5 & 79.2 & 79.2 & 42.5 & 85.0 \\
RP\textsubscript{EBO} & 88.2 & 80.4 & 80.4 & 82.1 & 85.0 \\
RP\textsubscript{GradNorm} & 18.7 & 11.2 & 11.2 & 50.0 & 85.0 \\
NCI & 87.7 & 79.7 & 79.7 & 81.0 & 85.0 \\
ConfBranch & 84.2 & 77.1 & 77.1 & 77.9 & 85.5 \\
OpenGAN &45.1  &  35.5    & 35.5  &50.0 & 82.2 \\
LogitNorm & 89.2 & 81.0 & 81.0 & 82.3 & 85.7 \\
    GODIN &  72.8 &   63.7 &    63.7 &  63.4 &  83.1 \\
                   RotPred &  86.1 &   78.2 &    78.2 &  79.4 &  85.3 \\
OE & 91.9 & 80.5 & 80.5 & 83.0 & 84.0 \\
UDG & 83.9 & 70.1 & 70.1 & 75.3 & 76.0 \\
MixOE & 88.1 & 79.7 & 79.7 & 81.4 & 84.9 \\
Energy & 91.7 & 80.4 & 80.4 & 82.9 & 83.8 \\
NovelBranch & 89.5 & 81.0 & 81.0 & 82.5 & 85.6 \\
Extended & 88.4 & 80.1 & 80.1 & 81.0 & 85.8 \\
\bottomrule
\end{tabular}
\end{small}
\end{center}
\end{table*}

\subsection{OSR performance when using ImageNet-1K pretrained weights}
We present the result in Table \ref{tab: main result with ImageNet-1K pretrained weights}. There is no result for OpenMax as it requires prediction of the test set to cover all training species. Though the ground truths of the test set cover all training species, the predictions of this model do not. 
\begin{table*}[!htbp]
\caption{\textbf{Benchmarking results on Open-Insect for C-America with ImageNet-1K pretrained weights.} We evaluate approaches falling into three categories: i) post-hoc methods, ii) training-time regularization (without auxiliary data), and iii) outlier exposure with auxiliary data. Results are shown for the C-America.  For each of the three open-set splits -- local (L), non-local (NL), and non-moth (NM) -- the AUROC is reported along with the accuracy of the closed-set test set. Since C-America is the dataset with the smallest training dataset, we also report results using ImageNet-1K pretrained weights compared to the models trained from scratch in Table~\ref{tab: main result}.}
\label{tab: main result with ImageNet-1K pretrained weights}
\begin{center}
\begin{small}

\begin{tabular}{l|ccc}
\toprule
 &   \multicolumn{3}{c}{C-America} \\ 
  & L & NL & closed-set Acc. \\
\midrule
\multicolumn{4}{l}{\textbf{Post-hoc methods} }\\
\rowcolor{Moccasin} OpenMax & \textcolor{lightgray}{N/A} & \textcolor{lightgray}{N/A} & \\
\rowcolor{Moccasin} MSP  & 86.01 & 87.02 & \\
\rowcolor{Moccasin} TempScale & 85.42 & 86.54 & \\
\rowcolor{Moccasin} ODIN  & 69.83 & 70.72 & \\
\rowcolor{Moccasin} MDS  & 86.95 & 88.39 & \\
\rowcolor{Moccasin} MDSEns  & 57.1 & 59.38 & \\
\rowcolor{Moccasin} RMDS  & \textbf{89.0} & \textbf{91.35} & \\
\rowcolor{Moccasin} Gram & 38.06 & 42.7 & \\
\rowcolor{Moccasin} EBO  & 86.48 & 88.72 & \\
\rowcolor{Moccasin} OpenGAN  & 64.41 & 68.17 & \\
\rowcolor{Moccasin} GradNorm  & 37.05 & 35.6 & \\
\rowcolor{Moccasin} ReAct & 86.62 & 88.79 & \\
\rowcolor{Moccasin} MLS  & 87.61 & 89.67 & \\
\rowcolor{Moccasin} KLM  & 87.36 & 88.5 & \\
\rowcolor{Moccasin} VIM & 88.05 & 89.74 & \\
\rowcolor{Moccasin} KNN  & 86.3 & 88.01 & \\
\rowcolor{Moccasin} DICE  & 19.12 & 16.5 & \\
\rowcolor{Moccasin} RankFeat  & 68.33 & 69.27 & \\
\rowcolor{Moccasin} ASH & 85.19 & 86.67 & \\
\rowcolor{Moccasin} SHE & 57.55 & 60.8 & \\
\rowcolor{Moccasin} NECO   & 80.87 & 81.76 & \\
\rowcolor{Moccasin} FDBD  & 87.76 & 88.97 & \\
\rowcolor{Moccasin} RP-MSP  & 86.0 & 87.15 & \\
\rowcolor{Moccasin} RP-ODIN  & 70.28 & 73.75 & \\
\rowcolor{Moccasin} RP-EBO & 85.35 & 88.09 & \\
\rowcolor{Moccasin} RP-GradNorm  & 27.62 & 25.73 &\multirow{-26}{*}{\textbf{89.14}} \\
\midrule
\multicolumn{4}{l}{\textbf{Training-time regularization} }\\
\rowcolor{LemonChiffon1} ConfBranch & 81.38 & 93.23 & 89.69 \\
\rowcolor{LemonChiffon1} LogitNorm   &\textbf{88.41}& \underline{\textbf{95.38}} & \textbf{89.86}\\
\rowcolor{LemonChiffon1} ARPL & 87.99 & 92.17 & 89.29 \\
\rowcolor{LemonChiffon1} G-ODIN & 71.02 & 65.93 & 88.42 \\
\rowcolor{LemonChiffon1} RotPred& 80.28 & 92.04 & 89.28 \\
\midrule
\multicolumn{4}{l}{\textbf{Training with auxiliary data} }\\
\rowcolor{AliceBlue} OE  &89.16 & 93.32 & 88.71 \\
\rowcolor{AliceBlue} MCD & 88.54 & 91.38 & 88.38 \\
\rowcolor{AliceBlue} UDG  &82.27 & 88.97 & 78.91 \\
\rowcolor{AliceBlue}MIXOE  & 90.07 & 94.24 & 89.12 \\
\rowcolor{AliceBlue} Energy  & 90.64 & 93.66 & 88.0 \\
\rowcolor{AliceBlue} NovelBranch & \underline{\textbf{92.05}} & \textbf{95.35} & 89.83 \\
\rowcolor{AliceBlue} Extended & 91.19 & 93.75 & \underline{\textbf{90.67}} \\
\bottomrule
\end{tabular}
\end{small}
\end{center}
\vskip -0.1in
\end{table*}

\begin{table*}[!htbp]
\caption{\textbf{Benchmarking results on Open-Insect with different training and post-hoc method combinations.} We present the results obtained from different combinations of training methods (with or without auxiliary training data) with selected post-hoc methods.
Results are shown for the three regions in Open-Insect: NE-America, W-Europe, C-America.  For each of the three open-set splits -- local (L), non-local (NL), and non-moth (NM) -- the AUROC is reported. The colored \colorbox{blue!30}{AUROC} are those shown in Table \ref{tab: main result}. }
\label{tab: full_result}
\vskip 0.15in
\begin{center}
\begin{small}
\resizebox{0.95\textwidth}{!}{%
\begin{tabular}{cc|ccc|ccc|ccc}
\toprule
 & & \multicolumn{3}{c|}{NE-America} &   \multicolumn{3}{c|}{W-Europe} &  \multicolumn{3}{c}{C-America} \\ 
  Training & Post-hoc&  L & NL & NM & L & NL & NM  & L & NL & NM \\
  \midrule
\midrule
\multicolumn{11}{l}{\textbf{Training regularization} }\\
   & MSP & 77.45 & 84.28 & 94.0 & 76.87 & 85.19 & 88.7 & 87.35 & 89.12 & 90.22 \\
   &EBO & 80.55 & 87.25 & 95.32 & 80.77 & 87.72 & 95.61 & 87.3 & 89.54 & 90.01 \\
   &MLS & \colorbox{blue!30}{80.58}  & \colorbox{blue!30}{87.29} & \colorbox{blue!30}{95.31} & \colorbox{blue!30}{80.78} & \colorbox{blue!30}{87.73} & \colorbox{blue!30}{95.59} & 87.27 & 89.44 & 90.15 \\
   \multirow{-4}{*}{LogitNorm} &TempScale& 79.96 & 86.5 & 94.99 & 79.2 & 86.06 & 94.87 & \colorbox{blue!30}{87.56} & \colorbox{blue!30}{89.5} & \colorbox{blue!30}{90.47} \\
\midrule
\multicolumn{11}{l}{\textbf{Training with auxiliary data} }\\
  & MSP & \colorbox{blue!30}{79.75} & \colorbox{blue!30}{86.33} & \colorbox{blue!30}{90.12} & 75.36 & 84.58 & 89.6 & 89.54 & 94.03 & 92.16 \\
  & MLS & 56.43 & 52.96 & 54.37 & 53.68 & 47.23 & 44.7 & 84.58 & 86.13 & 88.28 \\
  & EBO & 55.82 & 52.05 & 53.1 & 50.71 & 42.65 & 39.19 & 73.26 & 68.79 & 75.8 \\
 \multirow{-5}{*}{OE} & TempScale& 79.48 & 86.36 & 90.14 & \colorbox{blue!30}{75.38} & \colorbox{blue!30}{84.58} & \colorbox{blue!30}{89.61} & \colorbox{blue!30}{89.51} & \colorbox{blue!30}{94.04} & \colorbox{blue!30}{92.14} \\
\midrule
  & MSP & 74.11 & 80.22 & 92.56 & - & - & - & \colorbox{blue!30}{80.76} & \colorbox{blue!30}{83.08} & \colorbox{blue!30}{87.89} \\
  & MLS & \colorbox{blue!30}{75.1} & \colorbox{blue!30}{81.94} & \colorbox{blue!30}{91.48} & - & - & - & 80.36 & 81.75 & 85.9 \\
  & EBO & 71.73 & 77.45 & 75.45 & - & - & - & 72.1 & 70.65 & 70.9 \\
  \multirow{-5}{*}{UDG}  & TempScale& 74.63 & 80.19 & 92.1 & - & - & - & 80.51 & 82.31 & 87.34 \\
\midrule
 &  MSP & \colorbox{blue!30}{86.16} & \colorbox{blue!30}{92.45} & \colorbox{blue!30}{94.39} & \colorbox{blue!30}{85.21} & \colorbox{blue!30}{91.93} & \colorbox{blue!30}{94.1} & \colorbox{blue!30}{86.19} & \colorbox{blue!30}{87.94} & \colorbox{blue!30}{90.12} \\
  & MLS & 72.43 & 78.56 & 90.61 & 69.34 & 75.95 & 90.04 & 85.51 & 87.06 & 91.03 \\
  & EBO & 72.03 & 78.01 & 90.29 & 68.9 & 75.35 & 89.73 & 84.91 & 86.36 & 90.57 \\
  \multirow{-5}{*}{MixOE} & TempScale& 86.99 & 93.99 & 94.81 & 85.7 & 93.58 & 94.64 & 86.04 & 88.03 & 89.94 \\
\midrule
  & MSP & 81.86 & 86.97 & 93.58 & 80.81 & 84.46 & 93.3 & 86.32 & 88.33 & 90.13 \\
 & MLS & \colorbox{blue!30}{87.37} & \colorbox{blue!30}{95.12} & \colorbox{blue!30}{92.53} & \colorbox{blue!30}{84.64} & \colorbox{blue!30}{94.79} & \colorbox{blue!30}{89.53} & 89.91 & 93.62 & 91.49  \\
  & EBO & 87.21 & 94.98 & 91.6 & 84.34 & 94.6 & 88.05 & \colorbox{blue!30}{89.99} & \colorbox{blue!30}{93.79} & \colorbox{blue!30}{91.22} \\
 \multirow{-5}{*}{Energy}  &TempScale& 82.1 & 87.53 & 94.82 & 81.04 & 85.04 & 94.76 & 87.39 & 89.76 & 91.56 \\
\midrule
  & MSP & 82.5 & 87.74 & 93.5 & 81.16 & 86.15 & 93.02 & 86.54 & 88.08& 90.47 \\
  & MLS & \colorbox{blue!30}{85.51} & \colorbox{blue!30}{94.07} & \colorbox{blue!30}{89.98} & \colorbox{blue!30}{83.94} & \colorbox{blue!30}{93.76} & \colorbox{blue!30}{91.74} &  \colorbox{blue!30}{87.77} &  \colorbox{blue!30}{89.65} &  \colorbox{blue!30}{91.10} \\
  & EBO & 85.31 & 93.92 & 88.96 & 83.67 & 93.63 & 90.82 & 87.67 & 89.54 & 90.82 \\
   \multirow{-5}{*}{NovelBranch} &TempScale& 83.17 & 88.86 & 95.19 & 81.94 & 87.79 & 95.06 & 87.20 &  88.95 &  91.84 \\
\midrule
  & MSP & 82.4 & 87.76 & 93.39 & 80.98 & 85.89 & 92.82 & 85.78 & 86.88 & 90.34 \\
 & MLS &\colorbox{blue!30}{83.49} & \colorbox{blue!30}{92.15} & \colorbox{blue!30}{86.08} &\colorbox{blue!30} {82.63} & \colorbox{blue!30}{91.94} & \colorbox{blue!30}{89.38} & \colorbox{blue!30}{86.91} & \colorbox{blue!30}{89.03} & \colorbox{blue!30}{89.31} \\
 & EBO & 83.25 & 91.94 & 84.86 & 82.33 & 91.75 & 88.29 & 86.81 & 88.94 & 88.95 \\
 \multirow{-5}{*}{Extended} &TempScale& 83.09 & 88.89 & 95.11 & 81.78 & 87.48 & 94.89 & 86.37 & 87.55 & 91.81 \\
\bottomrule
\end{tabular}
}
\end{small}
\end{center}
\end{table*}

\begin{table*}[]
\caption{AUROC scores for post-hoc methods with high standard deviation across three training runs. We highlight the AUROC in \textcolor{red}{red} when the OSR scores for positive (open-set) samples are generally lower than those for negative (closed-set) samples.}
\label{tab: large sd method results}
\vskip 0.15in
\begin{center}
\begin{small}
\resizebox{0.95\textwidth}{!}{%
\begin{tabular}{c|cccc|cccc|cccc}
\toprule
  & \multicolumn{4}{c|}{L} &   \multicolumn{4}{c|}{NL} &  \multicolumn{4}{c}{NM} \\ 
Method &  Run 1 & Run 2 & Run 3 & Mean\textsubscript{(standard deviation)} &  Run 1 & Run 2 & Run 3 & Mean\textsubscript{(standard deviation)}  & Run 1 & Run 2 & Run 3 & Mean\textsubscript{(standard deviation)}  \\
\midrule
\multicolumn{2}{l}{\textbf{NE-America}}\\
KNN \cite{sun2022out}& 76.61& \textcolor{red}{39.49}& 76.90&64.33 \textsubscript{(17.57)}& 84.43& \textcolor{red}{38.24}& 84.32&68.99 \textsubscript{(21.75)}& 95.83& \textcolor{red}{20.41}& 94.80&70.35 \textsubscript{(35.32)}\\
DICE \cite{sun2022dice}& 67.47& \textcolor{red}{42.77}& 67.38&59.21 \textsubscript{(11.62)}& 74.46& \textcolor{red}{41.15}& 73.11&62.91 \textsubscript{(15.39)}& 94.32& \textcolor{red}{37.38}& 92.21&74.64 \textsubscript{(26.36)}\\
SHE \cite{zhang2022out}& 68.56& \textcolor{red}{38.17}& 68.67&58.47 \textsubscript{(14.35)}& 76.23& \textcolor{red}{36.87}& 75.62&62.91 \textsubscript{(18.41)}& 88.49& \textcolor{red}{24.75}& 89.19&67.47 \textsubscript{(30.22)}\\
\midrule
\multicolumn{2}{l}{\textbf{W-Europe}}\\
KNN \cite{sun2022out}& 76.00& \textcolor{red}{51.13}& 74.79&67.30 \textsubscript{(11.45)}& 85.15& \textcolor{red}{43.34}& 84.91&71.13 \textsubscript{(19.65)}& 95.20& \textcolor{red}{35.19}& 93.95&74.78 \textsubscript{(28.00)}\\
DICE \cite{sun2022dice}& 68.66& \textcolor{red}{42.54}& 60.84&57.35 \textsubscript{(10.95)}& 79.13& \textcolor{red}{42.03}& 65.79&62.32 \textsubscript{(15.35)}& 94.37& \textcolor{red}{48.36}& 92.51&78.41 \textsubscript{(21.26)}\\
SHE \cite{zhang2022out}& 67.13& \textcolor{red}{44.63}& 63.47&58.41 \textsubscript{(9.86)}& 73.65& \textcolor{red}{39.18}& 70.75&61.19 \textsubscript{(15.61)}& 87.83& \textcolor{red}{34.27}& 83.54&68.55 \textsubscript{(24.30)}\\

\bottomrule
\end{tabular}
}
\end{small}
\end{center}
\end{table*}

\subsection{Comparing the efficiency of post-hoc methods}
\label{sec: inference requirement}
All methods were evaluated with 1 RTX8000 GPU, 8 CPUs, 8 workers, and 100 GB CPU memory.  The classifier used is the ResNet50 C-America classifier. In Table \ref{tab: post-hoc efficiency}, if ``Data needed for setup'' is ``Train'', the setup time was calculated with 3,167 images, a subset of C-America training images.  If ``Data needed for setup'' is ``Val'', the setup time was calculated with 2,000 images, a subset of C-America validation images. Setup only need to be done once and the time needed is independent of the number of test images. Here, the inference time in Table \ref{tab: post-hoc efficiency}
is the time to process 4,000 test images (2,000 closed-set and 2,000 open-set).  Inference time increases as the number of test images increases. Some methods require hyperparameters. We indicate this by placing \checkmark in the ``Hyperparameter Search'' column of Table \ref{tab: post-hoc efficiency}.

\begin{table*}[!htbp]
\caption{\textbf{Comparison of post-hoc method efficiency.} We compare the efficiency of post-hoc methods in terms of setup time, access to training data, inference efficiency, and hyperparameter search requirements. Times are reported in seconds. }
\label{tab: post-hoc efficiency}
\begin{center}
\begin{small}
\begin{tabular}{lcccc}
\toprule
Method & \makecell{Setup time \\ (Sec.)} & \makecell{Data needed \\ for setup} & \makecell{Inference time \\ (Sec.)} & \makecell{Hyperparameter \\ search} \\
\midrule
OpenMax \cite{bendale2016towards} & 10.47 \textsubscript{(2.47)}&Train & 34.24 \textsubscript{( 2.82)} &\\
MSP \cite{hendrycks2016baseline} &\textcolor{lightgray}{N/A} &\textcolor{lightgray}{N/A} & 7.52 \textsubscript{( 1.98)} &\\
TempScale \cite{guo2017calibration} & 5.63 \textsubscript{(1.96)}&Val & 5.16 \textsubscript{( 0.86)} &\\
ODIN \cite{liang2017enhancing} &\textcolor{lightgray}{N/A} &\textcolor{lightgray}{N/A} & 11.05 \textsubscript{( 2.29)} &\checkmark\\
MDS \cite{lee2018simple} & 17.24 \textsubscript{(1.29)}&Train & 66.92 \textsubscript{( 7.89)} &\\
MDSEns \cite{lee2018simple} & 10.90 \textsubscript{(1.65)}&Train & 9.21 \textsubscript{( 0.24)} &\checkmark\\
RMDS \cite{ren2021simple} & 27.12 \textsubscript{(1.56)}&Train & 70.41 \textsubscript{( 2.22)} &\\
Gram \cite{pmlr-v119-sastry20a} & 19.12 \textsubscript{(1.64)}&Train & 17.19 \textsubscript{( 1.93)} &\checkmark\\
EBO \cite{liu2020energy} &\textcolor{lightgray}{N/A} &\textcolor{lightgray}{N/A} & 4.92 \textsubscript{( 0.49)} &\checkmark\\
GradNorm \cite{huang2021importance} &\textcolor{lightgray}{N/A} &\textcolor{lightgray}{N/A} & 10.91 \textsubscript{( 1.87)} &\\
ReAct \cite{sun2021react} & 6.12 \textsubscript{(1.67)}&Val & 5.65 \textsubscript{( 0.43)} &\checkmark\\
MLS \cite{hendrycks2019scaling} &\textcolor{lightgray}{N/A} &\textcolor{lightgray}{N/A} & 10.74 \textsubscript{( 1.77)} &\\
KLM \cite{pmlr-v162-hendrycks22a} & 7.71 \textsubscript{(2.12)}&Val & 1180.58 \textsubscript{( 9.89)} &\\
VIM \cite{wang2022vim} & 22.57 \textsubscript{(2.19)}&Train & 8.73 \textsubscript{( 2.01)} &\checkmark\\
KNN \cite{sun2022out} & 11.54 \textsubscript{(2.25)}&Train & 6.85 \textsubscript{( 0.93)} &\checkmark\\
DICE \cite{sun2022dice} & 10.46 \textsubscript{(1.65)}&Train & 6.12 \textsubscript{( 0.34)} &\\
RankFeat \cite{song2022rankfeat} &\textcolor{lightgray}{N/A} &\textcolor{lightgray}{N/A} & 19.40 \textsubscript{( 1.87)} &\\
ASH \cite{djurisic2022extremely} &\textcolor{lightgray}{N/A} &\textcolor{lightgray}{N/A} & 6.10 \textsubscript{( 1.22)} &\checkmark\\
SHE \cite{zhang2022out} & 10.71 \textsubscript{(2.13)}&Train & 5.87 \textsubscript{( 0.72)} &\\
NECO \cite{ammar2023neco} & 16.85 \textsubscript{(3.71)}&Train & 6.74 \textsubscript{( 2.02)} &\checkmark\\
FDBD \cite{liu2023fast} & 10.30 \textsubscript{(1.55)}&Train & 5.97 \textsubscript{( 1.30)} &\checkmark\\
RP\textsubscript{MSP} \cite{jiang2023detecting} &\textcolor{lightgray}{N/A} &Train & 9.88 \textsubscript{( 3.46)} &\\
RP\textsubscript{ODIN}  \cite{jiang2023detecting} &\textcolor{lightgray}{N/A} &Train & 9.90 \textsubscript{( 0.30)} &\checkmark\\
RP\textsubscript{EBO}  \cite{jiang2023detecting} &\textcolor{lightgray}{N/A} &Train & 4.93 \textsubscript{( 0.42)} &\checkmark\\
RP\textsubscript{GradNorm}  \cite{jiang2023detecting} &\textcolor{lightgray}{N/A} &Train & 9.94 \textsubscript{( 3.16)} &\\
NCI \cite{liu2023detecting} & 9.53 \textsubscript{(1.88)}&Train & 5.18 \textsubscript{( 0.75)} &\checkmark\\
\bottomrule
\end{tabular}
\end{small}
\end{center}
\end{table*}

\begin{figure}
\begin{center}
\centerline{\includegraphics[width=0.8\textwidth]{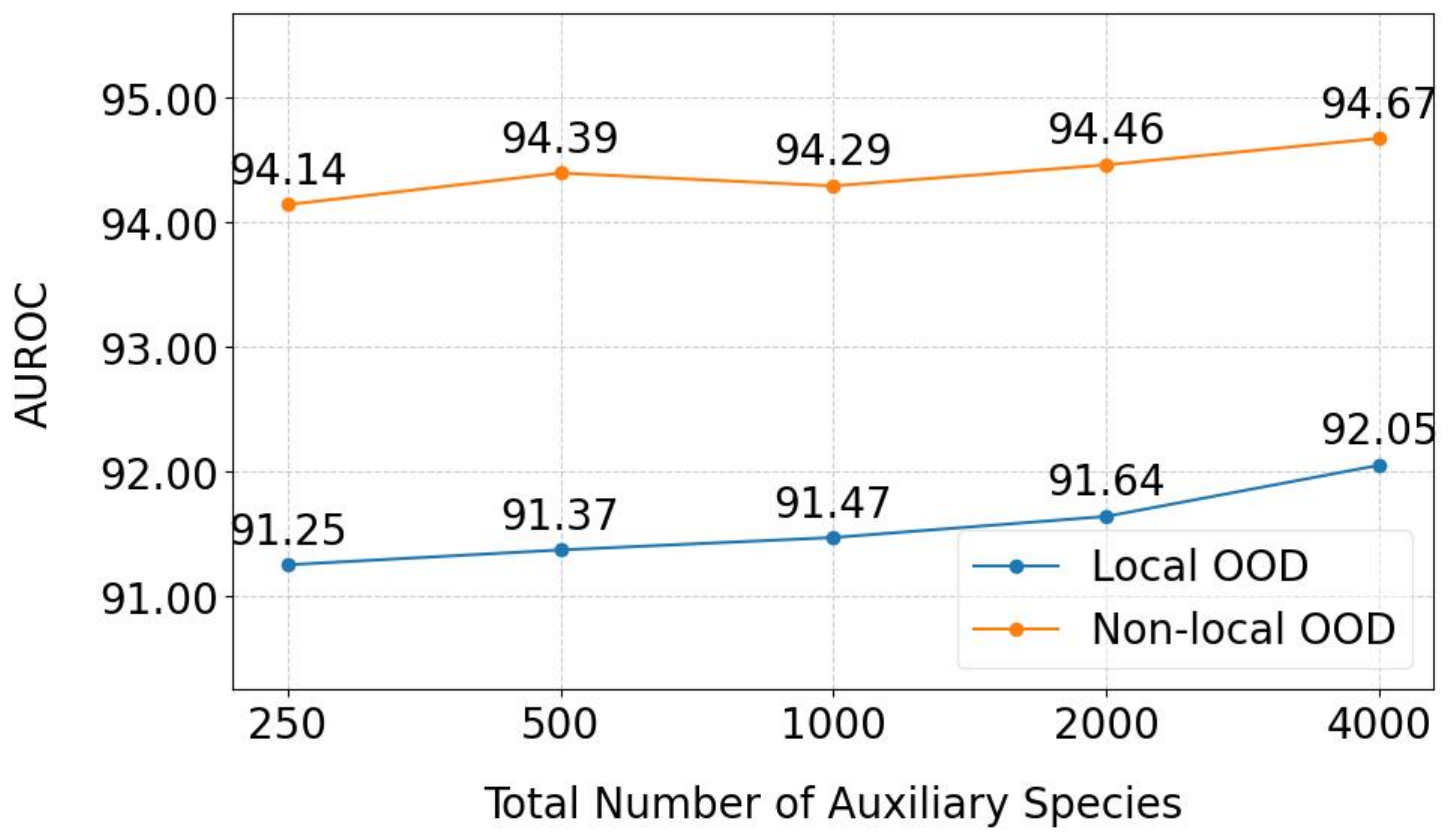}}
\caption{\textbf{Species diversity of the auxiliary dataset.} We vary the number of species in the auxiliary dataset, while keeping the total number of images fixed to 80,000.}
\label{fig:diversity ablation}
\end{center}
\end{figure}

\section{Explainability of OSR methods}

We constructed the following dataset to empirically verify that \textit{background features are not enough to achieve good OSR performance on Open-Insect}. We first used Grounding DINO \cite{liu2024grounding} to detect and generate a bounding box around the insect in each image, then used it as an input prompt to Segment Anything (SAM) \cite{kirillov2023segment}, which segmented the insect. Finally, we replaced the segmented region with the average value of the surrounding pixels. We applied this pipeline to a subset of Open-Insect C-America. Since both object detection and segmentation were done by models, errors can occur. Hence, we manually verified a (randomized) subset of the processed images and discarded the rest, ultimately obtaining 859 valid closed-set and 1433 valid local open-set images for our experiments. We show an example  in Fig. \ref{fig:teaser_boxplot_examples}. 

\begin{figure}[h]
    \centering
    \begin{subfigure}[b]{0.3\textwidth}
        \centering
        \includegraphics[width=\textwidth]{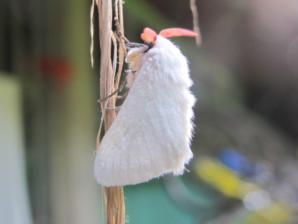}
        \caption{Original image}
    \end{subfigure}
    \hfill
    \begin{subfigure}[b]{0.3\textwidth}
        \centering
        \includegraphics[width=\textwidth]{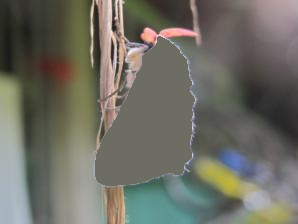}
        \caption{Moth masked}
    \end{subfigure}
    \hfill
    \begin{subfigure}[b]{0.3\textwidth}
        \centering
        \includegraphics[width=\textwidth]{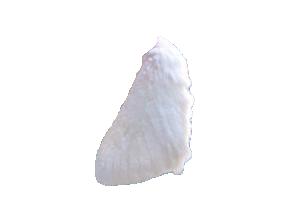}
        \caption{Background masked}
    \end{subfigure}
     \caption{A mask generated by SAM which misses legs and antennae.}
   \label{fig:parts_missing}
\end{figure}

We believe the slight drop of performance when the background is masked can be explained by the quality of the machine-generated masks. During verification, we noticed that the masks generated by SAM often miss legs and antennae (see Fig. \ref{fig:parts_missing}). As a result, when masking the background, parts of the insect such as legs and antennae were also inadvertently masked, which may explain the slight drop in performance compared to the original images.

\begin{figure}[h]

    \centering
    \begin{subfigure}[b]{0.3\textwidth}
        \centering
        \includegraphics[width=\textwidth]{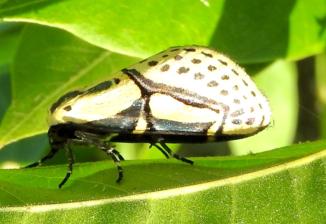}
        \caption{Original image}
    \end{subfigure}
    \hfill
    \begin{subfigure}[b]{0.3\textwidth}
        \centering
        \includegraphics[width=\textwidth]{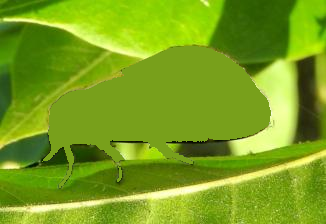}
        \caption{Moth masked}
    \end{subfigure}
    \hfill
    \begin{subfigure}[b]{0.3\textwidth}
        \centering
        \includegraphics[width=\textwidth]{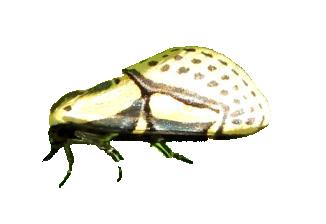}
        \caption{Background masked}
    \end{subfigure}
    \caption{An example from the subset we used for the explainability experiment. The mask was automatically generated by SAM and the color of the mask is the average value of the surrounding pixels.}
   \label{fig:explainability_image_example}
\end{figure}

\section{Performance of BioCLIP on Open-Insect}
\label{sec:bioclip_experiments}

BioCLIP \cite{stevens2024bioclip}, a foundation model for species recognition, is increasingly used for biodiversity-related tasks. Hence, we conducted additional experiments to evaluate the OSR performance obtained directly from BioCLIP as well as models finetuned from BioCLIP vision encoder weights.

\subsection{Off-the-shelf performance of BioCLIP on Open-Insect}

Since not all closed-set species of Open-Insect are included in BioCLIP’s training set and some open-set species overlap with BioCLIP’s training data, we evaluated BioCLIP’s off-the-shelf performance on a subset of Open-Insect. We include closed-set species that are included in Tree-of-Life 10M (training data of BioCLIP) \cite{stevens2024bioclip} or "seen" by BioCLIP and open-set species that are ``unseen'' to evaluate its OSR performance. We list the number of species of each regional split in Table \ref{tab:bioclip_zero_shot_subsets}. We also include 4,633 ``unseen'' non-local moth species in this subset.

\begin{table}[!htbp]
\caption{\textbf{Number of species in each region of the Open-Insect subset.} Seen closed-set species appear in both the Open-Insect and BioCLIP training sets, while unseen open-set species are open-set species in Open-Insect and do not appear in BioCLIP training set.}
\label{tab:bioclip_zero_shot_subsets}
\begin{center}
\begin{tabular}{lrrr}
\toprule
Category &	NE-America	& W-Europe	& C-America \\
\midrule
Seen closed-set &	1260 &	177	& 71\\
Unseen open-set &	418	&447	&1537\\
\bottomrule
\end{tabular}
\end{center}
\vskip -0.1in
\end{table}

We compared BioCLIP’s performance on this subset with our ResNet50 classifiers trained from scratch. We used three simple post-hoc methods, MSP, MLS, and EBO. We find that open-set recognition on Open-Insect is challenging for BioCLIP, likely because Lepidoptera (moths and butterflies) represent less than 2\% of the species in its training data (Table~\ref{tab:bioclip_zero_shot_result}).

\begin{table*}[!htbp]

\caption{\textbf{Comparison of BioCLIP’s performance on the Open-Insect subset with our ResNet50 classifiers trained from scratch.} }
\label{tab:bioclip_zero_shot_result}
\begin{center}
\begin{small}
\resizebox{\textwidth}{!}{%
\begin{tabular}{cc|rrr|rrr|rrr}
\toprule
\multirow{ 2}{*}{Model} & \multirow{ 2}{*}{Post-hoc Method} & \multicolumn{3}{c|}{NE-America} &   \multicolumn{3}{c|}{W-Europe} &  \multicolumn{3}{c}{C-America} \\ 
  & & L & NL  & closed-set Acc. & L & NL  & closed-set Acc. & L & NL & closed-set Acc. \\
  \midrule

\multirow{3}{*}{BioCLIP} & MSP & 65.24	& 60.38 & \multirow{3}{*}{24.37} & 68.83&69.30&\multirow{3}{*}{37.56}&74.83&	73.33&\multirow{3}{*}{55.75}\\
  & MLS & 77.86 &	72.96 & & 76.44&	79.54&&79.64	&85.72\\
  & EBO & 55.94 &	56.19 &  & 56.05&	57.10&&53.08	&63.73\\
  \midrule
  \multirow{3}{*}{Ours} & MSP & 87.71	& 93.97 &  \multirow{3}{*}{90.49} &85.61&	92.74 &\multirow{3}{*}{89.07}&83.89	&84.52&\multirow{3}{*}{78.52}\\
  & MLS & 75.25 &	76.96 &&69.64	&75.49&&74.63&	74.61\\
  & EBO & 74.60 &	75.93 && 68.99&	74.57&&71.89&	71.71\\
  \bottomrule
\end{tabular}
}
\end{small}
\end{center}
\end{table*}

\subsection{Using BioCLIP pretrained weights}

It has been shown that using ImageNet pretrained weights can help closed-set accuracy \cite{van2021benchmarking} and that there is a positive correlation between the closed-set and open-set performance \cite{vaze2021open}. The results we show in Table \ref{tab: main result with ImageNet-1K pretrained weights} show that those observations hold for Open-Insect as well. Since BioCLIP was trained on more domain specific data, we finetuned two ViT-B-16 models \cite{dosovitskiy2021an} for C-America - one with BioCLIP pretrained weights, and the other with ImageNet-1K pretrained weights, to compare the effect of different pretraining data. 

Both models were fine-tuned for 30 epochs with all 636 C-America closed-set species with 224 $\times$ 224 images. We only finetuned the last block and kept other parameters frozen. The OSR performance was evaluated with all 636 closed-set species, 1537 “unseen” local (L) open-set, and 4,633 “unseen” non-local (NL) open-set species. We present the results in Table \ref{tab:finetuned_vit_result}. We find that using BioCLIP weights is more effective than ImageNet-1K.


\end{document}